\documentclass{article}

\PassOptionsToPackage{numbers,compress}{natbib}
\usepackage[eandd, preprint]{neurips_2026}

\usepackage[utf8]{inputenc}
\usepackage[T1]{fontenc}
\usepackage{hyperref}
\usepackage{url}
\usepackage{booktabs}
\usepackage{amsfonts}
\usepackage{nicefrac}
\usepackage{microtype}
\usepackage{xcolor}
\usepackage{amsmath}
\usepackage{multirow}
\usepackage{subcaption}
\usepackage{wrapfig}
\usepackage[capitalise,noabbrev]{cleveref}
\usepackage{makecell}

\newcommand{\benchmark}{\textsc{SciConvBench}}

\usepackage[most]{tcolorbox}
\tcbuselibrary{listings,breakable,skins}

\newtcblisting{promptbox}[1]{
  enhanced,
  breakable,
  colback=blue!5,
  colframe=black!20,
  boxrule=0.4pt,
  arc=2pt,
  left=4pt,
  right=4pt,
  top=4pt,
  bottom=4pt,
  title={#1},
  fonttitle=\bfseries,
  listing only,
  listing options={
    basicstyle=\ttfamily\small,
    breaklines=true,
    columns=fullflexible
  }
}

\title{\benchmark{}: Benchmarking LLMs on Multi-Turn Clarification for Task Formulation in Computational Science}

\author{%
  Nithin Somasekharan \\
  Rensselaer Polytechnic Institute \\
  Troy, NY \\
  \texttt{somasn@rpi.edu}     
  \And
  Youssef Hassan \\
  Rensselaer Polytechnic Institute \\
  Troy, NY \\
  \texttt{hassay@rpi.edu}    
  \And
  Shiyao Lin \\
  University of Texas at Arlington \\
  Arlington, TX \\
  \texttt{shiyao.lin@uta.edu}
  \And
  Gihan Panapitiya \\
  Pacific Northwest National Laboratory \\
  Richland, WA \\
  \texttt{gihan.panapitiya@pnnl.gov}
  \AND
  Patrick Emami \\
  National Renewable Energy Laboratory \\
  Golden, CO \\
  \texttt{Patrick.Emami@nrel.gov}
  \And
  Anurag Acharya \\
  Pacific Northwest National Laboratory \\
  Richland, WA \\
  \texttt{anurag.acharya@pnnl.gov}    
  \And
  Sameera Horawalavithana \\
  Pacific Northwest National Laboratory \\
  Richland, WA \\
  \texttt{sameera.horawalavithana@pnnl.gov}    
  \And
  Shaowu Pan\thanks{Corresponding author.} \\
  Rensselaer Polytechnic Institute \\
  Troy, NY \\
  \texttt{pans2@rpi.edu}
}

\begin{document}

\maketitle

\begin{abstract}
Large Language Models (LLMs) are increasingly deployed as scientific AI assistants, and a growing body of benchmarks evaluates their capabilities across knowledge retrieval, reasoning, code generation, and tool use.
These evaluations, however, typically assume the scientific problem is already well-posed, whereas practical scientific assistance often begins with an ill-posed user request that must be refined through dialogue before any computation, analysis, or experiment can be carried out reliably. 
We introduce \benchmark{}, a benchmark for multi-turn clarification in scientific task formulation across four computational science problem domains: \emph{fluid mechanics}, \emph{solid mechanics}, \emph{materials science}, and \emph{partial differential equations (PDEs)}. 
\benchmark{} targets two complementary capabilities: eliciting missing information (\emph{disambiguation}) and detecting and correcting erroneous requests containing internally contradictory information (\emph{inconsistency resolution}). 
Our benchmark pairs a structured task ontology with a rubric-based evaluation framework, enabling systematic measurement of LLM performance across three dimensions: clarification behavior, conversational grounding, and final-specification fidelity.
Current frontier models perform relatively well on \emph{inconsistency resolution}, but even the best model resolves only $52.7\%$ of the disambiguation cases in \emph{fluid mechanics}. 
We further find that frontier LLMs frequently make silent assumptions and perform implicit specification repairs that are not grounded in the conversation with users. 
\benchmark{} establishes a foundation for evaluating the upstream conversational reasoning that a reliable computational science assistant requires. The code and data can be found at \url{https://github.com/csml-rpi/SciConvBench}.
\end{abstract}

\section{Introduction}

\begin{wrapfigure}{r}{0.38\textwidth}
  \vspace{-2.4em}
  \centering
  \includegraphics[width=0.38\textwidth]{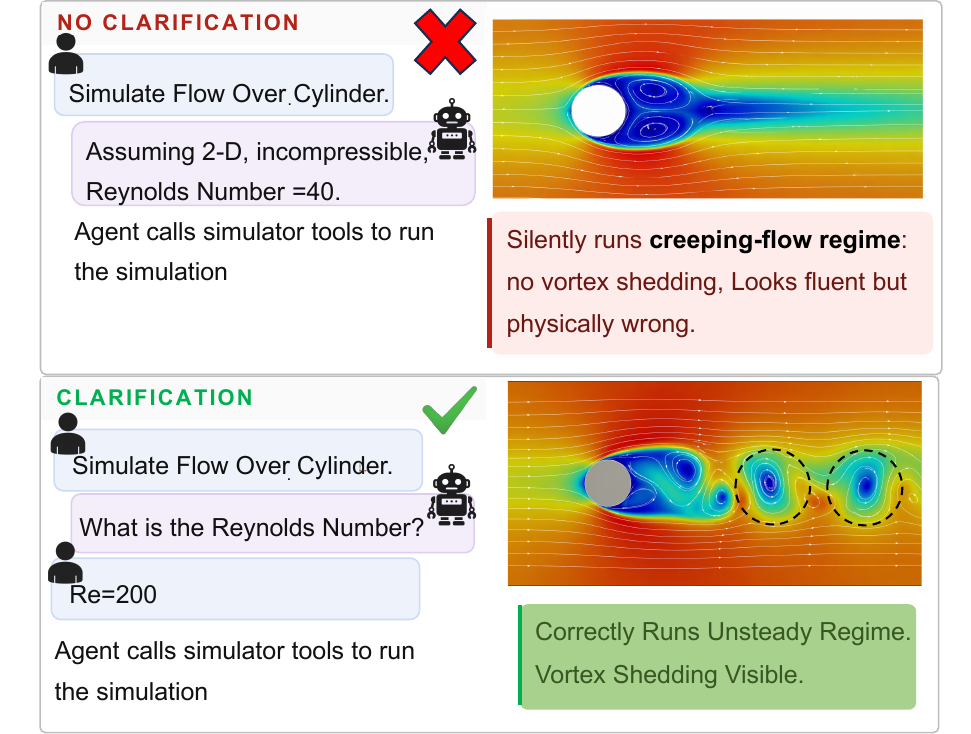}
  \caption{\small Flow over a cylinder showing how skipped clarification leads to a wrong flow regime.}
  \label{fig:clarification_need}
  \vspace{-1.0em}
\end{wrapfigure}

Large language models (LLMs) are increasingly used as conversational interfaces for computational science, supporting scientific question answering~\cite{somasekharan2026cfdllmbench}, code generation~\cite{tian2024scicode}, and agentic execution of scientific simulation workflows~\cite{yue2025foam,pandey2025openfoamgpt}. Yet most scientific  benchmarks for LLMs assess these capabilities given complete problem formulation, typically assuming a clean task statement with fixed objectives, constraints, and expected outputs \cite{wang2024scibench,sun2024scieval,tian2024scicode,liu2025mattools,chen2024scienceagentbench}. This omits an upstream failure mode in scientific practice: before a model can compute, write code, or invoke tools reliably, it may first need to transform an incomplete or internally inconsistent user request into a well-specified scientific task. 
In computational science, such formulation errors are consequential because a missing boundary condition, ambiguous material property, incompatible constitutive assumption, missing Reynolds number, or contradictory numerical constraint can alter the underlying problem, yielding a specification that is physically invalid, irreproducible, or misaligned with the user's intent. For example, \cref{fig:clarification_need} illustrates the downstream consequence of unresolved prompt issues: if a critical parameter such as the Reynolds number is not clarified, an agent may silently run a plausible but incorrect flow regime, wasting computation and producing a result that is irrelevant to the intended scientific task. 

\begin{wraptable}{r}{0.42\linewidth}
  \vspace{-0.8em}
  \centering
\centering
\scriptsize
\setlength{\tabcolsep}{2.5pt}
\renewcommand{\arraystretch}{1.2}
\begin{tabular}{llcc}
\toprule
\textbf{Benchmark} & \textbf{Domain} & \textbf{$n$} & \textbf{\shortstack[l]{Resolution \\ Rate}} \\
\midrule
CLAMBER \cite{zhang2024clamber} & General            & 115 & 86.1\,\% \\
\midrule
\multirow{4}{*}{\shortstack[l]{\textsc{SciConvBench}\\(ours)}}
                                & Fluid mechanics    & 151 & 18.2\,\% \\
                                & Solid mechanics    & 228 & 29.4\,\% \\
                                & Materials science  & 130 & 53.8\,\% \\
                                & PDEs               &  61 & 65.6\,\% \\
\bottomrule
\end{tabular}
  \vspace{-0.8em}
\label{tab:clamber_vs_sciconvbench}
\caption{Comparison between a filtered subset of CLAMBER~\citep{zhang2024clamber} (a general-domain clarification dataset) and the disambiguation split of \benchmark{}, both evaluated with \textsc{Gemini 2.5 Pro}. Resolution rate drops drastically on \benchmark{}, indicating that computational-science domains pose a substantially harder clarification challenge than general-domain prompts.}
\end{wraptable}
Existing clarification and ambiguity benchmarks study follow-up questioning mainly in general-purpose or information-seeking settings \cite{lee2023clarificationqa,zhang2024clamber,aliannejadi2019qulac,kuhn2023clam,gan2024clarqllm}, while multi-turn and agent benchmarks show that information distributed across dialogue remains difficult for current models \cite{deshpande2025multichallenge,kwan2024mteval,yao2025taubench,laban2026lost}. These benchmarks, however, do not reflect the kinds of clarifications computational science actually demands, and the gap is quantitative as well as conceptual: under the same model and protocol, \textsc{Gemini~2.5~Pro} resolves $86.1\%$ of cases on a filtered subset of CLAMBER but drops to as low as $18.2\%$ on \textsc{SciConvBench} disambiguation (\cref{tab:clamber_vs_sciconvbench}). The missing evaluation setting is whether a model can identify missing or conflicting scientific requirements and resolve them through dialogue before producing a final task specification. These results motivate a science-specific clarification benchmark, since general clarification benchmarks do not stress the domain groundedness required in computational science.

\begin{figure}[h]
  \centering
  \includegraphics[width=\textwidth]{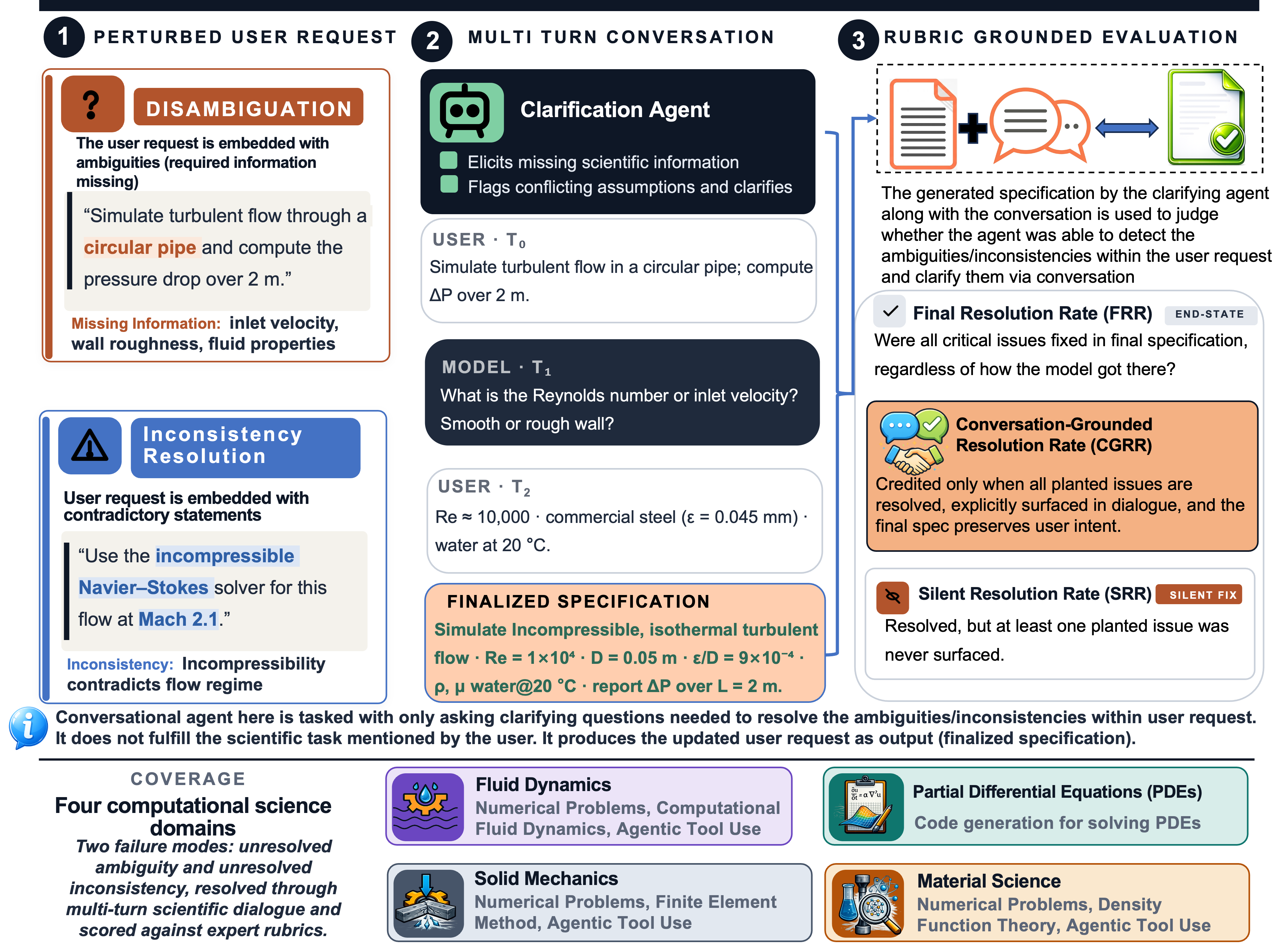}
  \caption{Overview of \benchmark{}. The benchmark spans four computational science domains and two task types. For each instance, a model interacts with a simulated user to resolve missing or conflicting information and then produces a final specification. Evaluation compares the final specification against the reference specification while using the full conversation as context to assess whether the model resolved the ambiguity or inconsistency in the original scientific request.}
  \label{fig:overview}
  \vspace{-1em}
\end{figure}

We introduce \benchmark{} (\cref{fig:overview}), a benchmark for multi-turn clarification of scientific task formulation across domains such as \emph{fluid mechanics}, \emph{solid mechanics}, \emph{materials science}, and \emph{partial differential equations (PDEs)}. Each instance begins with a scientific request containing either missing information, which requires disambiguation, or conflicting information, which requires inconsistency resolution. The model interacts with a user over multiple turns and then produces a final clarified specification. \benchmark{} evaluates \emph{conversational scientific task formulation}, defined as the ability of a model to resolve incomplete or internally inconsistent scientific requests through dialogue and produce a usable final prompt or specification. 


Our goal is to shift evaluation upstream. Before asking whether a model can solve, code, or execute a scientific task, we ask whether it can help define the task correctly. This paper makes three contributions. First, we formalize conversational scientific task formulation as a benchmark setting centered on unresolved ambiguity and unresolved inconsistency. Second, we introduce an evaluation framework that separates intent faithful final resolution from conversation-grounded resolution, exposing silent assumptions and silent repairs that standard end-state metrics can miss. Third, we benchmark current models across scientific domains and ontology categories, and analyze the robustness of conclusions across judges, prompts, and user simulators. Our results show that this upstream stage remains difficult for frontier models: no single model dominates across all tasks and domains, inconsistency resolution is substantially easier than missing-information elicitation, the leading model changes across the two tasks, and every model exhibits a persistent gap between final correctness and conversation-grounded resolution, with a larger gap on inconsistency resolution tasks than on disambiguation. 


\section{Related work}

\paragraph{Clarification and ambiguity.}
Clarifying-question research has long studied when an assistant should ask rather than answer. Conversational retrieval and QA benchmarks such as \textsc{Qulac}, \textsc{ClariQ}, and \textsc{ClarQ} evaluate clarification selection, ranking, generation, and large-scale question mining~\citep{aliannejadi2019qulac,aliannejadi2020convai3,kumar2020clarq}; \textsc{AmbigQA} and \textsc{CAmbigNQ} treat ambiguous questions as requiring multiple interpretations or explicit clarification before answering~\citep{min2020ambigqa,lee2023clarificationqa}; and \textsc{CLAMBER}, \textsc{CondAmbigQA}, \textsc{CLAM}, \textsc{Apa}, future-turn RLHF, and proactive information-gathering work formalize ambiguity taxonomies, conditional ambiguity, strategic clarification, and high-value question asking under incomplete context~\citep{zhang2024clamber,li2025condambigqa,kuhn2023clam,kim2024apa,zhang2025futureturn,huang2025proactive}. Related inconsistency and rule-grounded dialogue benchmarks include \textsc{CONTRADOC}, which localizes document contradictions, and \textsc{ShARC}, which requires follow-up questions when rule-grounded requests are underspecified~\citep{li2024contradoc,saeidi2018sharc}. \textsc{QuestBench} isolates information gathering for missing logical or mathematical preconditions, and \textsc{ClarQ-LLM} shows LLMs often answer instead of clarifying in task-oriented dialogue~\citep{fu2025questbench,gan2024clarqllm}. These benchmarks establish clarification as a measurable capability, but their ambiguities are primarily about which sense of a polysemous query is meant, which subtopic of a search the user cares about, which of several valid factoid readings to return, or which user preference to follow, rather than tied to scientific regimes.
\paragraph{Multi-turn, agentic, and simulator-based evaluation.}
Multi-turn evaluation has moved from general dialogue quality to interaction robustness. \textsc{MT-Bench} and LLM-as-a-judge evaluation exposed both the scalability and biases of automatic multi-turn judging~\citep{zheng2023judging}; \textsc{Chatbot Arena}, \textsc{Arena-Hard-Auto}, and length-controlled AlpacaEval study human-aligned large-scale ranking and verbosity control~\citep{chiang2024chatbot,li2024arenahard,dubois2024alpacaeval}; and \textsc{MT-Eval}, \textsc{MultiChallenge}, \textsc{LLMs Get Lost}, and \textsc{RMTBench} show that models remain brittle when evidence is distributed across turns or users behave less cooperatively~\citep{kwan2024mteval,deshpande2025multichallenge,laban2026lost,xiang2025rmtbench}. Agent benchmarks such as \textsc{AgentBench}, \textsc{WebArena}, \textsc{GAIA}, \textsc{SWE-bench}, \textsc{MINT}, and $\tau$-bench evaluate tool, API, website, codebase, or simulated-user environments~\citep{liu2023agentbench,zhou2024webarena,mialon2024gaia,jimenez2024swebench,wang2024mint,yao2025taubench}. Because these settings increasingly rely on user simulators, recent work studies simulator fidelity and robustness: $\tau$-bench and $\tau^2$-Bench adopt LLM-simulated users for scalable agent evaluation~\citep{yao2025taubench,barres2025tau2}; \textsc{MirrorBench}, reliable-simulator work, \textsc{SimulatorArena}, and non-collaborative simulators analyze when simulated users preserve or distort measured assistant quality~\citep{hathidara2026mirrorbench,luo2024reliableusersim,dou2025simulatorarena,shim2026noncollaborative}; and broader task-oriented and social-simulation work provides context for this protocol~\citep{davidson2023usersim,ni2026usersimsurvey,argyle2023silicon,budzianowski2018multiwoz}.

\paragraph{Scientific benchmarks and domain-specific agents.}
Scientific evaluation has advanced rapidly, but most benchmarks assume that the task is already specified. \textsc{SciBench} and \textsc{SciEval} evaluate scientific reasoning and research tasks~\citep{wang2024scibench,sun2024scieval}; \textsc{SciCode}, \textsc{MatTools}, \textsc{ScienceAgentBench}, \textsc{SciAgent}, and \textsc{ChemCrow} evaluate research coding, materials-science tool use, data-driven discovery, tool-augmented scientific reasoning, and chemistry agents~\citep{tian2024scicode,liu2025mattools,chen2024scienceagentbench,ma2024sciagent,bran2024chemcrow}. Computational-science agents and benchmarks similarly target executable workflows after formulation: \textsc{OpenFOAMGPT}, \textsc{NL2FOAM}, \textsc{CFDLLMBench}, and \textsc{MetaOpenFOAM} for fluids and CFD~\citep{pandey2025openfoamgpt,dong2025nl2foam,somasekharan2025cfdllmbench,chen2024metaopenfoam}; \textsc{FEABench}, \textsc{AutoFEA}, and \textsc{ALL-FEM} for solids, FEA, PDE formulation, and code generation~\citep{mudur2025feabench,hou2025autofea,deotale2026allfem}; and \textsc{HoneyComb} and \textsc{MechAgents} for materials and mechanics workflows~\citep{zhang2024honeycomb,ni2024mechagents}. Our focus is the preceding conversational step: whether the model elicits or flags the scientific commitments needed to make execution meaningful.

\section{SciConvBench}
\label{sec:sciconvbench}

\subsection{Benchmark Scope and Domains}

The benchmark spans four computational-science domains: \emph{fluid mechanics}, \emph{solid mechanics}, \emph{materials science}, and \emph{partial differential equations (PDEs)} and includes both general numerical problem statements and prompts requiring the invocation of domain-specific simulator tool. Each domain covers a different class of scientific task formulation (see Equation~\ref{eq:task-def}). \emph{Fluid Mechanics} includes general fluid-mechanics problems and Computational Fluid Dynamics (CFD) prompts. \emph{Solid Mechanics} includes mechanics and finite-element-style task formulation. \emph{Materials Science} includes materials-science reasoning and Density Functional Theory (DFT) based task formulations. \emph{Partial Differential Equations (PDEs)} includes mathematical PDE problem specification and numerical setup tasks.

\subsection{Task Definition}

We define a scientific task formulation as a structured specification of the physical or computational study to be performed~\citep{ashton2025fluidintelligenceforwardlook}. A clean task $y^\star$ is written as
\begin{equation}
y^\star :=
\big[
\nu_{\mathrm{obj}},
\nu_{\mathrm{geom}},
\nu_{\mathrm{model}},
\nu_{\mathrm{prop}},
\nu_{\mathrm{bc}},
\nu_{\mathrm{ic}},
\nu_{\mathrm{num}},
\nu_{\mathrm{out}},
\nu_{\mathrm{tool}}
\big],
\label{eq:task-def}
\end{equation}
where the entries denote the objective of the study, geometry or computational domain, governing physics or constitutive model, material or transport properties, boundary conditions, initial conditions, numerical controls, requested outputs, and tool-specific settings, collectively defining the \emph{ontology} of a scientific task. A benchmark instance is obtained by perturbing a clean task $y^\star$ into an initial user request $x=T_z(y^\star)$. The perturbation set $z=\{(k_j,\tau_j)\}_{j=1}^{m}$ records the planted issues, where $k_j$ indexes one of the entries of \cref{eq:task-def} and $\tau_j\in\{\textsc{missing},\textsc{conflict}\}$. If $\tau_j=\textsc{missing}$, information required by entry $k_j$ is omitted or left underspecified in the initial request. If $\tau_j=\textsc{conflict}$, the request contains mutually incompatible information for that entry, or an incompatibility between that entry and another part of the specification. The model interacts with the user over multiple turns and finally produces a specification $\hat{y}$. The benchmark evaluates whether $\hat{y}$ resolves all planted issues, preserves the intended task, and reaches this resolution through conversation rather than silent guessing or unannounced correction.

\subsection{Interaction Protocol}

Each interaction begins from the transformed user request. The conversational agent may ask clarification questions over multiple turns before producing its final output. The agent is instructed to ask only one question per turn. The user responds only from the hidden reference specification for that instance and does not provide information outside the intended task. To keep interactions comparable across models and domains, we use a fixed turn budget of 11. This choice follows directly from dataset construction: each instance contains at most 10 planted ambiguities or inconsistencies, so 11 turns are sufficient in principle to address all issues in a case and produce a final specification. The conversation terminates either when the model explicitly finalizes the task or when the turn limit is reached (\cref{app:forced_finilization,app:turn-cap}), at which point the model must produce its final clarified specification. \textsc{SciConvBench} does not require solver execution, code execution, or tool invocation for scoring; prompts may be tool-oriented, but evaluation is restricted to conversational task formulation, allowing the benchmark to be evaluated by any conversational agentic framework.

\subsection{Dataset Creation}

We construct \textsc{SciConvBench} in two stages. We first collect a pool of clean, well-posed scientific tasks, and then manually convert them into conversational instances containing either missing information (\emph{disambiguation}) or conflicting information (\emph{inconsistency resolution}). This design ensures that every benchmark item starts from a scientifically valid reference problem, and that the difficulty comes from task formulation rather than from noisy or ill-posed source data.

\paragraph*{Source pool}
We assemble source tasks from vetted educational, benchmark, and tool-informed resources across four computational-science domains. Fluid and PDE tasks draw from standard texts, FoamBench and CFDCodeBench within CFDLLMBench, and \textsc{SciCode}~\citep{munson2020fundamentals,white2021fluid,somasekharan2025cfdllmbench,tian2024scicode}. Solid mechanics tasks use standard mechanics texts and finite-element resources, including \textsc{FEABench}, \textsc{AutoFEA}, \textsc{ALL-FEM}, FEniCS, and CalculiX~\citep{beer2021mechanics,goodno2018mechanics,ugural2021advanced,mudur2025feabench,hou2025autofea,deotale2026allfem,logg2012fenics,dhondt1998calculix}. Materials tasks combine textbook problems \citep{callister2018materials,askeland2025science,shackelford2021materials} with DFT tasks drawing from \textsc{MaScQA}, \textsc{MatSciBench}, and \textsc{MatTools}~\citep{zaki2024mascqa,zhang2025matscibench,liu2025mattools}.

\paragraph*{Prompt transformation}
Each source item is first normalized into a clean reference prompt admitting a coherent scientific answer or setup. For \emph{disambiguation} cases, we remove information that a responsible assistant should request before finalizing the task, such as boundary conditions, constitutive assumptions, material or transport properties, solver settings, geometry details, target outputs, or numerical tolerances. For \emph{inconsistency} cases, we insert incompatible or conflicting statements while keeping the overall request realistic. This transformation is performed prompt-by-prompt rather than through automatic templates, since the missing or conflicting information is strongly domain- and problem-dependent. Each missing entity or planted inconsistency is also tagged to one of the components in \cref{eq:task-def}.

\paragraph*{Expert review and filtering}

\begin{wrapfigure}{r}{0.30\textwidth}
  \vspace{-1.2em}
  \centering
  \includegraphics[width=0.30\textwidth]{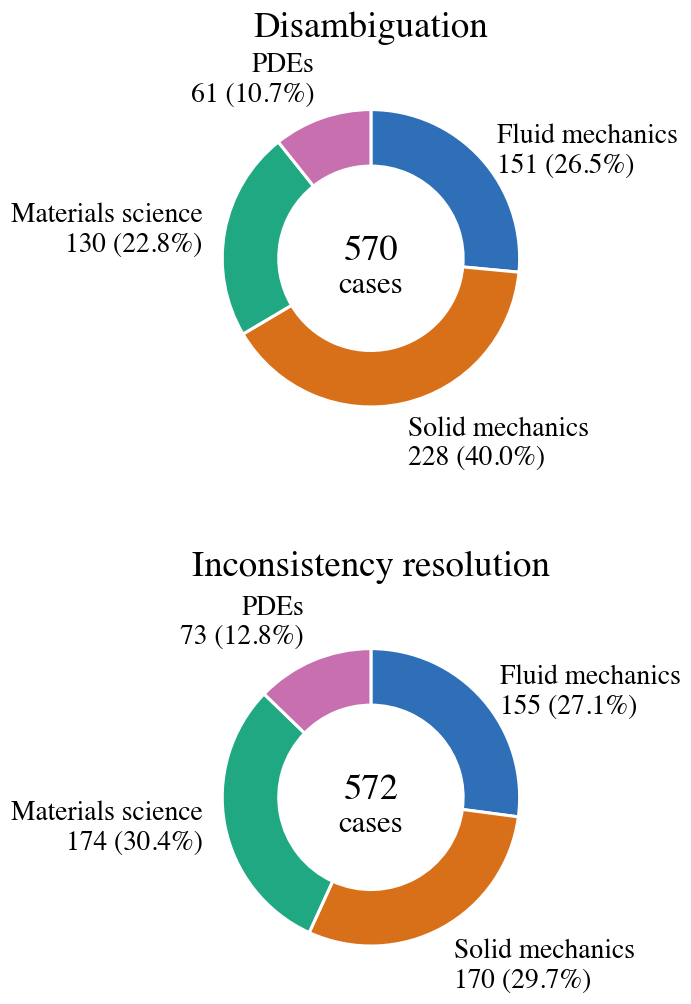}
  \caption{Case distribution across the four \benchmark{} domains.}
  \label{fig:dataset-stats}
  \vspace{-3em}
\end{wrapfigure}

Quality control was performed by experts who were not involved in authoring the original transformed prompt. Reviewers checked that the hidden or conflicting information was scientifically meaningful, that the case admitted a clear intended resolution, that the prompt did not leak the answer through trivial cues, and that the conversational variant remained realistic. After pilot benchmarking, we removed cases that were too trivial, too underdetermined, or solved uniformly well across all tested models. The final case split across 1,142 total cases  is shown in \cref{fig:dataset-stats}.


\subsection{Evaluation Protocol}
Following recent conversational benchmark design~\citep{bai2024mtbench101,deshpande2025multichallenge,yao2025taubench}, we separate final output success from conversation-grounded success, since a model may guess or silently repair missing scientific details without resolving them through dialogue. Each instance is evaluated as a structured judgment problem using the conversation transcript, the final specification, and the reference issue annotation. Because correct resolutions can vary in wording and dialogue path, exact string matching and handwritten heuristics are insufficient. We therefore use an LLM judge with an expert-curated rubric that defines the planted issue per case, successful resolution criteria, and the evidence required for conversational grounding. For every case, the judge is supplied with that case's specific missing entities or planted inconsistencies. This protocol follows prior evidence that strong LLM judges can reach high agreement with humans on open-ended evaluation when guided by explicit rubrics~\citep{zheng2023judging,singhal2025profbench,hashemi2024llmrubric}. 

\subsection{Metrics}
\label{sec:metrics}

\paragraph{Case-level Rates.}\label{par:case_metric}
We evaluate whether a model turns an incomplete or inconsistent scientific request into a correct final task specification. For each case $i$, we compute three binary checks. \textbf{1) Resolution} ($R_i$): a binary metric that takes a value of 1 when all annotated issues of the case are resolved in the final specification produced by the agent, and 0 otherwise; \textbf{2) Conversational Grounding} ($G_i$): a binary metric that takes a value of 1 when all annotated issues of the case are explicitly clarified with the user, and 0 otherwise; and \textbf{3) Intent Fidelity} ($I_i$): a binary metric that takes a value of 1 when the final specification preserves the user's intended scientific task, and 0 otherwise. We combine these three checks into the case-level rates. \textbf{a) Final Resolution Rate} (FRR $\uparrow$): FRR measures the fraction of cases where the final specification is correct ($R_i=1$) and intent-faithful ($I_i=1$), regardless of whether the model reached that specification through dialogue ($G_i$ may be either 0 or 1). Higher FRR means that the model resolves more cases in its final output. \textbf{b) Conversation-Grounded Resolution Rate} (CGRR $\uparrow$):
CGRR measures the fraction of cases where the final specification is correct ($R_i=1$), intent-faithful ($I_i=1$), and grounded in the conversation ($G_i=1$). Higher CGRR means that the model resolves more cases through explicit clarification rather than silent guessing. \textbf{c) Silent Resolution Rate} (SRR $\downarrow$):
SRR measures the fraction of cases where the final specification is correct ($R_i=1$) and intent-faithful ($I_i=1$), but not grounded in the conversation ($G_i=0$). These cases correspond to implicit assumptions or silent repairs. Lower SRR is better because it means fewer cases are resolved without being made explicit to the user.

\paragraph{Component-level Rates.} \label{par:component_metric}
Unlike the case-level rates above, which credit a case only when all of its issues are resolved, component-level rates score each annotated issue separately. This gives a direct readout of which ontology components in \cref{eq:task-def} are resolved, grounded, or silently repaired. For issue $j$ in case $i$, let $k_{ij}\in y^\star$ be its ontology component. Let $r_{ij}=1$ if the issue is resolved in the final specification, and let $g_{ij}=1$ if the issue is clarified with the user; otherwise these values are 0. We also require the final specification to preserve the user's intent, i.e., $I_i=1$. Let $\mathcal{I}_{k}=\{(i,j):k_{ij}=k\}$ denote all issues belonging to component $k$. We define component-level
\textbf{1) Final Resolution Rate} (FRR$(k)$ $\uparrow$):
the fraction of issues in component $k$ that are resolved in an intent-faithful final specification. \textbf{2) Conversation-Grounded Resolution Rate} (CGRR$(k)$ $\uparrow$):
the fraction of issues in component $k$ that are both resolved and clarified with the user. \textbf{3) Silent Resolution Rate} (SRR$(k)$ $\downarrow$):
the fraction of issues in component $k$ that are resolved but not clarified with the user. 

\paragraph{Capability, Robustness, and Usability.}
CGRR gives the primary success criterion, but it does not explain why a model succeeds or fails. We therefore also report three diagnostic axes for the Pareto analysis: \emph{Capability}, \emph{Robustness}, and \emph{Usability}. Each axis is computed as an equally weighted average of lower-level diagnostic metrics. \emph{Capability} measures whether the model asks the right clarification questions and produces a complete final specification; it averages \emph{clarification recall}, the fraction of annotated issues surfaced by the model, \emph{clarification precision}, the fraction of the model's questions that target annotated issues, and \emph{plan completeness}, the fraction of required fields correctly instantiated in the final specification. \emph{Robustness} measures whether the model avoids unreliable dialogue behavior; it averages \emph{assumption avoidance}, \emph{error detection}, and \emph{memory consistency}, capturing silent assumptions, silent repairs, and contradictions with information established during the dialogue. \emph{Usability} measures whether the final specification remains aligned with the user's intended scientific task, using \emph{intent capture}. These axes are used only as diagnostic summaries; the main success metric remains CGRR. Full definitions and aggregation details are provided in \cref{sec:pareto_axis_define}.

\section{Experimental Setup}
\label{sec:experimental-setup}

\paragraph{Conversational evaluation framework.}
Each instance is evaluated as a multi-turn interaction between a conversational assistant model and a user. The assistant receives an ambiguous or inconsistent request, may ask one clarification question per turn, and produces a final specification either after explicit finalization or at the turn-budget cap. In the primary \emph{guided} setting, the assistant is instructed to act as a requirements analyst that identifies missing information, detects contradictions, and clarifies one issue at a time before finalizing. We evaluate five guided-mode models across all four domains: \textsc{Claude Sonnet 4.6}\cite{anthropic2026claude46}, \textsc{Gemini 2.5 Pro}\cite{google2025gemini25}, \textsc{Gemini 2.5 Flash}\cite{google2025gemini25}, \textsc{GPT-5.2}\cite{openai2025gpt52}, and \textsc{GPT-OSS-120B}\cite{openai2025gptoss}. Token usage per model and cost are detailed in \cref{app:compute-cost}. We also evaluate an \emph{unguided} control without explicit instructions to identify inconsistencies or ambiguities, since prior work shows that LLMs often fail to detect such issues unless prompted to do so~\citep{tyen-etal-2024-llms}; the guided-vs.-unguided comparison for \textsc{Gemini 2.5 Pro} is reported in \cref{app:guided-vs-unguided}. The user answering the question is a simulator LLM. The user simulator has access to the incomplete request, a hidden reference request specification, dialogue history, and it is strictly instructed to answer only from the reference request specification, using ``make a reasonable assumption'' when the requested detail is absent in the reference. In this way the simulator LLM abstains from providing information from its own knowledge and strictly adheres to the provided reference specification which is ambiguity or inconsistency free. This user simulator uses \textsc{claude sonnet 4.6} as the model, unless specified otherwise. Our use of LLM simulating user follows recent conversational-agent benchmarks using LLM user proxies~\citep{yao2025taubench,barres2025tau2,hathidara2026mirrorbench,luo2024reliableusersim,ni2026usersimsurvey}. Prompts used in assistant and simulator are provided in \cref{app:prompt-templates}.

\paragraph{Judge and human validation.}
All metrics are computed from saved conversations and final specifications using the rubric-based judging protocol in \cref{sec:sciconvbench}, which evaluates semantic resolution, conversational grounding, and intent preservation rather than exact string match. Unless otherwise noted, headline numbers use \textsc{Gemini 2.5 Pro} as the judge. To assess judge dependence, we rescore stratified subsets with alternative judges and compare automatic judgments against an 80-case human-annotated subset balanced by domain, task type, and outcome bucket. Human rater uses the same per-case rubric as the LLM judge, while blinded to the LLM-judge scores. Details are provided in \cref{app:human-annotation}.

\paragraph{Ablations on user simulator and agent prompt.}
Since both simulator choice and prompt wording can shift measured rates~\citep{dou2025simulatorarena,shim2026noncollaborative}, we report two ablations alongside the headline results. The \emph{simulator ablation} re-runs the 80-case stratified subset with the assistant fixed at \textsc{Gemini 2.5 Pro} and the user simulator varied across \textsc{Gemini 2.5 Pro}, \textsc{GPT-5.2}, and \textsc{Claude Sonnet 4.6}. The \emph{prompt-paraphrase ablation} fixes both the assistant at \textsc{Gemini 2.5 Pro} and replaces the guided-mode assistant prompt with two scientifically equivalent paraphrases that preserve the four behavioral contracts but vary role framing and instruction wording. Further details can be found in \cref{app:prompt-ablation,app:simulator-ablation}.

\section{Results}
\label{sec:results}

\subsection{Models, gaps, and the FRR--CGRR decomposition}

\begin{figure*}[h]
    \centering
    \includegraphics[width=\textwidth]{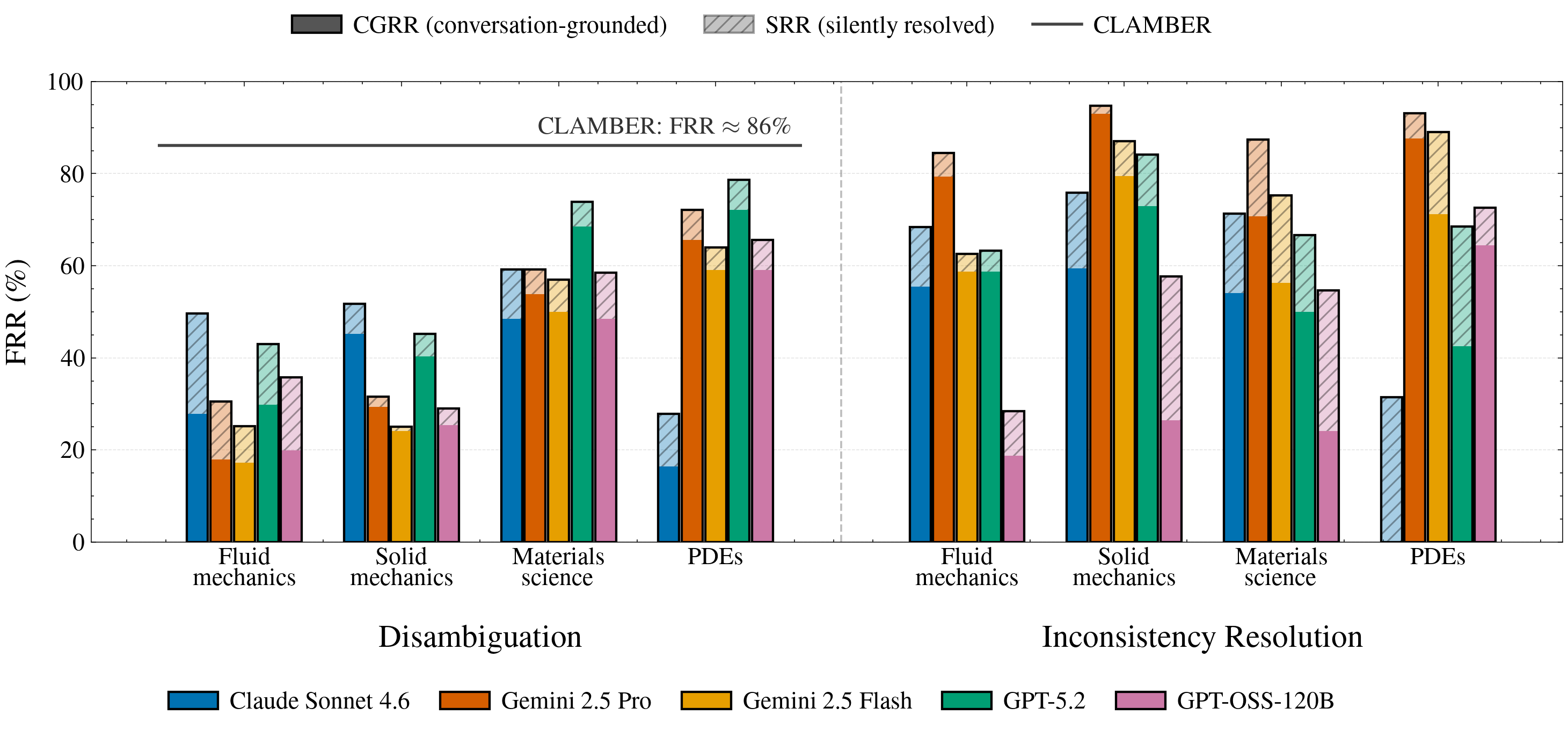}
    \caption{Case level resolution rate (\cref{par:case_metric}) comparison among different models for the different domains and tasks in \benchmark{}. FRR is further broken down into CGRR and SRR. The dotted horizontal line in the disambiguation block marks CLAMBER's reported FRR on general disambiguation tasks ($\approx 86\%$)~\citep{zhang2024clamber}.\benchmark{} is more challenging than general domain disambiguation benchmark like CLAMBER.}
    \label{fig:main-outcomes}
\end{figure*}

\Cref{fig:main-outcomes} decomposes Final Resolution Rate (FRR) into Conversation-Grounded Resolution Rate (CGRR) and Silent Resolution Rate (SRR), with full per-domain metrics deferred to \cref{tab:appendix_disambiguation,tab:appendix_inconsistency}. Every model exhibits a non-zero FRR – CGRR gap across domains and task types, averaging $8.2$ percentage points on Disambiguation and $14.7$ percentage points on Inconsistency Resolution (across the five evaluated LLMs). The leading models also differ by task: \textsc{GPT-5.2} is strongest on disambiguation, while \textsc{Gemini 2.5 Pro} is strongest on inconsistency resolution. This suggests that eliciting missing information and explicitly identifying contradictions are related but not identical capabilities. Fluid mechanics is consistently difficult for disambiguation. Inconsistency resolution shows a different pattern: the strongest models handle solids and PDE conflicts more reliably, while materials remains harder. Additional results can be found in \cref{app:additional-results}. 
\vspace{-0.6em}

\begin{figure*}[h]
    \centering
    \includegraphics[width=\textwidth]{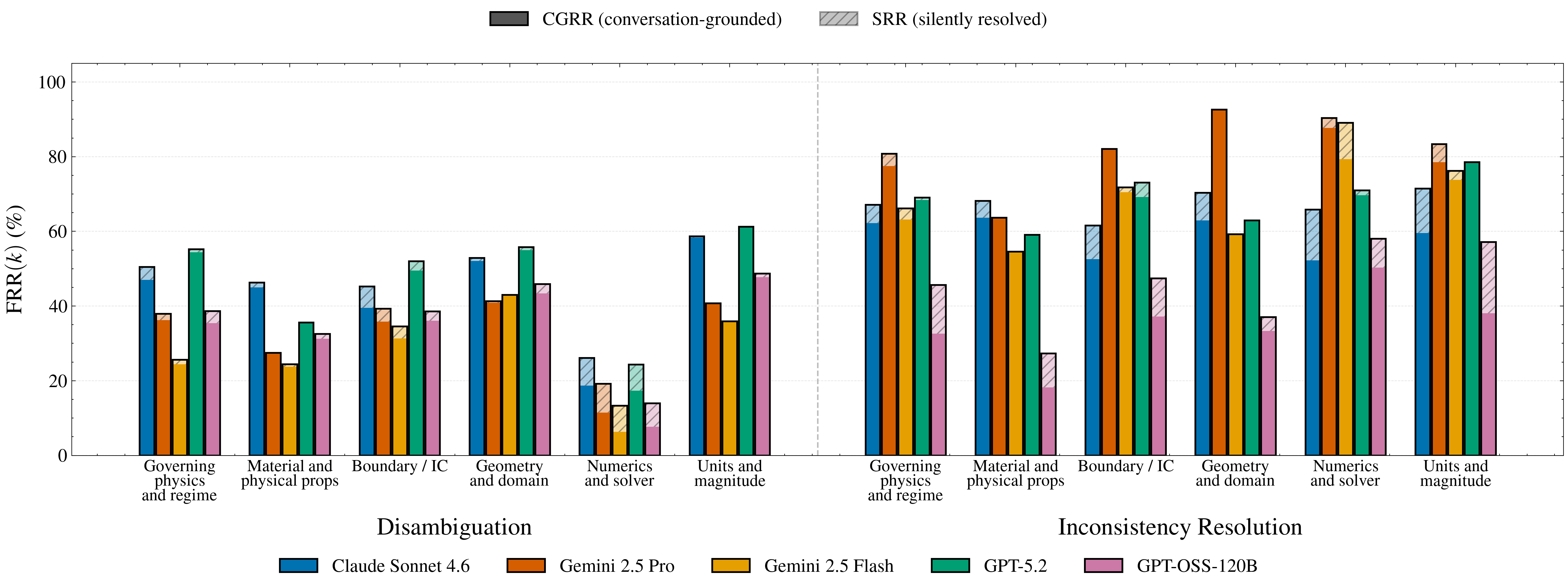}
    \caption{Component level resolution rate (\cref{par:component_metric}) comparison among different models for the different domains and tasks in \benchmark{} as defined in \cref{eq:task-def}. FRR is further broken down into CGRR and SRR. Not all components are equally challenging.}
    \label{fig:ontology-outcomes}
\end{figure*}

\paragraph{Ontology patterns.}
\Cref{fig:ontology-outcomes} reports component-level $\mathrm{FRR}(k)$, $\mathrm{CGRR}(k)$, $\mathrm{SRR}(k)$ as defined in Component-level Metrics in \cref{sec:metrics}, across the scientific components defined in \cref{eq:task-def}. Numerics and solver choices and governing-physics assumptions are the most fragile components, with the lowest component-wise FRR in the benchmark. Complete breakdown provided in \cref{app:ontology-breakdown}.
\vspace{-1em}

\begin{figure*}[h]
  \centering
  \includegraphics[width=\textwidth]{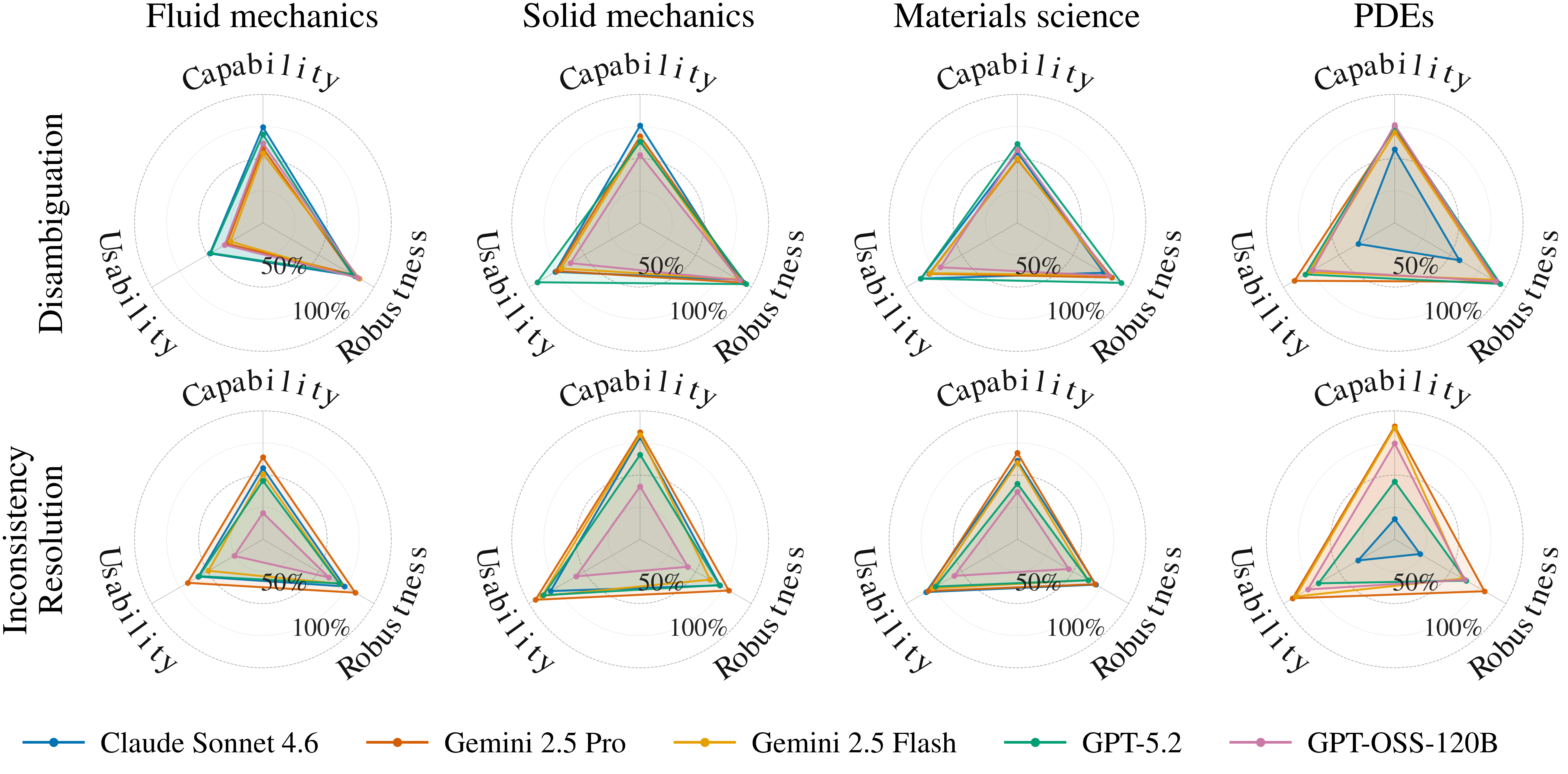}
  \caption{Pareto analysis across Capability, Robustness, and Usability. Top row: disambiguation. Bottom row: inconsistency resolution. Each panel is one domain; each trace is one model. Higher values are better.}
  \label{fig:pareto}
\end{figure*}

\paragraph{Pareto view.}
\Cref{fig:pareto} separates performance into Capability, Robustness, and Usability as defined in \cref{sec:metrics}. \textsc{GPT-5.2} has the strongest disambiguation profile, whereas \textsc{Gemini 2.5 Pro} is more robust on inconsistency resolution. Across models, Robustness is the least stable axis, especially when moving from missing-information cases to planted conflicts.

\vspace{-0.6em}

\paragraph{Robustness across judges, prompts, and simulators.}
\Cref{tab:robustness-summary} summarizes robustness checks on the 80-case subset. The conclusions are stable across judge choice, guided-prompt paraphrase, and user-simulator model.

\begin{table}[h]
\caption{Robustness checks on the 80-case stratified subset. Judge LLM choice, prompt paraphrase, and user-simulator LLM variation produce the same qualitative conclusions.}
\centering
\small
\setlength{\tabcolsep}{4pt}
\begin{subtable}[t]{0.32\linewidth}
\centering
\caption{Judge-human agreement (\%)}
\label{tab:robustness-judge}
\begin{tabular}{lcc}
\toprule
Judge & FRR & CGRR \\
\midrule
\textsc{Gemini 2.5 Pro} & 87.5 & 71.2 \\
\textsc{GPT-5.2}        & 87.5 & 71.2 \\
\textsc{Sonnet 4.6}     & 87.5 & 76.2 \\
\bottomrule
\end{tabular}
\end{subtable}\hfill
\begin{subtable}[t]{0.32\linewidth}
\centering
\caption{Prompt paraphrases (\%)}
\label{tab:robustness-prompt}
\begin{tabular}{lcc}
\toprule
Variant & FRR & CGRR \\
\midrule
Original  & 77.5 & 42.5 \\
Variant A & 75.0 & 45.0 \\
Variant B & 72.5 & 46.2 \\
\bottomrule
\end{tabular}
\end{subtable}\hfill
\begin{subtable}[t]{0.32\linewidth}
\centering
\caption{User simulators (\%)}
\label{tab:robustness-simulator}
\begin{tabular}{lcc}
\toprule
Simulator & FRR & CGRR \\
\midrule
\textsc{Gemini 2.5 Pro} & 72.5 & 46.2 \\
\textsc{GPT-5.2}        & 78.8 & 46.2 \\
\textsc{Sonnet 4.6}     & 77.5 & 42.5 \\
\bottomrule
\end{tabular}
\end{subtable}
\vspace{0.5em}
\label{tab:robustness-summary}
\vspace{-2em}
\end{table}
\section{Discussion}

\paragraph{No single model dominates both clarification regimes.} \textsc{GPT-5.2} is strongest on disambiguation, with CGRR of $52.7\%$, compared with $41.7\%$ for \textsc{Gemini 2.5 Pro}. However, the ordering reverses for inconsistency resolution: \textsc{Gemini 2.5 Pro} reaches $82.7\%$ CGRR, ahead of \textsc{Gemini 2.5 Flash} at $66.4\%$ and \textsc{GPT-5.2} at $56.0\%$. Thus, clarification ability is not a single scalar capability; eliciting missing information and confronting contradictions are separable scientific dialogue skills. Qualitative examples of grounded and silent resolutions are provided in \cref{app:qualitative-cases}.

\paragraph{Ontology-level structure explains where clarification breaks down.} Fluid mechanics is the clearest disambiguation bottleneck: the best fluid disambiguation CGRR is only $29.8\%$, while the best materials science and PDE disambiguation scores reach $68.5\%$ and $72.1\%$, respectively. The ontology analysis helps explain this gap. Numerics and solver choices are consistently weak, with component-level FRR only around $10\%$ to $21\%$ across models, and governing physics or regime assumptions are also fragile. These are not peripheral details; they determine what scientific problem is being solved. The benchmark therefore exposes failures in eliciting the commitments that make a scientific task well posed.

\paragraph{The Pareto analysis shows why no model should be treated as uniformly reliable.} Many traces are lopsided: models often preserve usability and the user's broad intent while losing robustness. For example, \textsc{GPT-5.2} has high disambiguation robustness, roughly $87\%$ to $94\%$ across domains, but its inconsistency robustness drops to roughly $61\%$ to $69\%$. \textsc{Gemini 2.5 Pro} also contracts relative to disambiguation, but remains the most balanced inconsistency model, combining high CGRR with strong capability and usability. This suggests that current models can often produce usable scientific specifications, but still fail to reliably surface conflicts or missing assumptions before finalizing the task.

\paragraph{The FRR/CGRR gap should be interpreted as silent scientific inference.} In many cases, the final response contains a plausible repair or default that was never explicitly elicited from the user. One extreme example is \textsc{Claude Sonnet 4.6} on PDE inconsistency, where FRR is $31.5\%$ but CGRR is $0.0\%$, meaning successful final repairs in that slice occur without guided clarification. This behavior can look intelligent: the model fills in a solver choice, physical regime, boundary convention, or material assumption. But because the assumption is not asked about or acknowledged, it cannot be audited. In scientific workflows, this is a reproducibility risk rather than merely an interaction flaw.
\vspace{-1em}
\section{Conclusion}
We introduced \textsc{SciConvBench}, a benchmark for conversational scientific
task formulation across fluid mechanics, solid mechanics, materials science, and PDEs. Unlike scientific LLM benchmarks that grade models after the task is specified, \textsc{SciConvBench} tests whether models surface missing or inconsistent scientific requirements through dialogue before committing to a final specification. Across five guided models, final resolution rate exceeds conversation-grounded resolution rate in every domain on both tasks, and the difference is stable across judges, prompt paraphrases, and simulators. Case-level evidence narrows the diagnosis: on the same planted issues, leading models reach the same correct specification but differ in whether they flag the conflict, so silent resolution is a confrontation failure, not a knowledge failure. By making this distinction measurable, \textsc{SciConvBench} establishes conversational task formulation as a necessary upstream capability for reliable scientific assistants.

\section{Acknowledgments and Disclosure of Funding}
This work was authored in part by the National Laboratory of the Rockies for the U.S. Department of Energy (DOE), operated under Contract No. DE-AC36-08GO28308. It was also supported in part by the Pacific Northwest National Laboratory, which is operated by Battelle Memorial Institute for the U.S. Department of Energy under Contract DE-AC05–76RLO1830. This material is based upon work supported by the U.S. Department of Energy, Office of Science, ASCR under Award Number DE-SC0025425. Any opinions, findings, and conclusions or recommendations expressed in this material are those of the author(s) and do not necessarily reflect the views of the DOE, the United States Government or any agency thereof. The U.S. Government retains and the publisher, by accepting the article for publication, acknowledges that the U.S. Government retains a nonexclusive, paid-up, irrevocable, worldwide license to publish or reproduce the published form of this work, or allow others to do so, for U.S. Government purposes. This paper has been cleared by PNNL for public release as PNNL-SA-222980. Shaowu Pan is supported by the Google Research Scholar Program. Computing resources are
supported by the Lambda research grant program, NSF-ACCESS-PHY240112 and by the National
Energy Research Scientific Computing Center under award NERSC ASCR-ERCAP0038273.

{\small
\bibliographystyle{plainnat}
\bibliography{references}

@inproceedings{lee2023clarificationqa,
  title={Asking Clarification Questions to Handle Ambiguity in Open-Domain QA},
  title = "Asking Clarification Questions to Handle Ambiguity in Open-Domain {QA}",
    author = "Lee, Dongryeol  and
      Kim, Segwang  and
      Lee, Minwoo  and
      Lee, Hwanhee  and
      Park, Joonsuk  and
      Lee, Sang-Woo  and
      Jung, Kyomin",
    editor = "Bouamor, Houda  and
      Pino, Juan  and
      Bali, Kalika",
    booktitle = "Findings of the Association for Computational Linguistics: EMNLP 2023",
    month = dec,
    year = "2023",
    address = "Singapore",
    publisher = "Association for Computational Linguistics",
    url = "https://aclanthology.org/2023.findings-emnlp.772/",
    doi = "10.18653/v1/2023.findings-emnlp.772",
    pages = "11526--11544",
}

@inproceedings{zhang2024clamber,
  title = "{CLAMBER}: A Benchmark of Identifying and Clarifying Ambiguous Information Needs in Large Language Models",
    author = "Zhang, Tong  and
      Qin, Peixin  and
      Deng, Yang  and
      Huang, Chen  and
      Lei, Wenqiang  and
      Liu, Junhong  and
      Jin, Dingnan  and
      Liang, Hongru  and
      Chua, Tat-Seng",
    editor = "Ku, Lun-Wei  and
      Martins, Andre  and
      Srikumar, Vivek",
    booktitle = "Proceedings of the 62nd Annual Meeting of the Association for Computational Linguistics (Volume 1: Long Papers)",
    month = aug,
    year = "2024",
    address = "Bangkok, Thailand",
    publisher = "Association for Computational Linguistics",
    url = "https://aclanthology.org/2024.acl-long.578/",
    doi = "10.18653/v1/2024.acl-long.578",
    pages = "10746--10766",
    }

@inproceedings{li2025condambigqa,
  title = "{C}ond{A}mbig{QA}: A Benchmark and Dataset for Conditional Ambiguous Question Answering",
    author = "Li, Zongxi  and
      Li, Yang  and
      Xie, Haoran  and
      Qin, S. Joe",
    editor = "Christodoulopoulos, Christos  and
      Chakraborty, Tanmoy  and
      Rose, Carolyn  and
      Peng, Violet",
    booktitle = "Proceedings of the 2025 Conference on Empirical Methods in Natural Language Processing",
    month = nov,
    year = "2025",
    address = "Suzhou, China",
    publisher = "Association for Computational Linguistics",
    url = "https://aclanthology.org/2025.emnlp-main.115/",
    doi = "10.18653/v1/2025.emnlp-main.115",
    pages = "2269--2288",
    ISBN = "979-8-89176-332-6",
}

@inproceedings{huang2025proactive,
  title = "Teaching Language Models To Gather Information Proactively",
    author = "Huang, Tenghao  and
      Chen, Sihao  and
      Chen, Muhao  and
      May, Jonathan  and
      Yang, Longqi  and
      Wan, Mengting  and
      Zhou, Pei",
    editor = "Christodoulopoulos, Christos  and
      Chakraborty, Tanmoy  and
      Rose, Carolyn  and
      Peng, Violet",
    booktitle = "Findings of the Association for Computational Linguistics: EMNLP 2025",
    month = nov,
    year = "2025",
    address = "Suzhou, China",
    publisher = "Association for Computational Linguistics",
    url = "https://aclanthology.org/2025.findings-emnlp.843/",
    doi = "10.18653/v1/2025.findings-emnlp.843",
    pages = "15588--15599",
    ISBN = "979-8-89176-335-7",
}

@inproceedings{aliannejadi2019qulac,
  author = {Aliannejadi, Mohammad and Zamani, Hamed and Crestani, Fabio and Croft, W. Bruce},
title = {Asking Clarifying Questions in Open-Domain Information-Seeking Conversations},
year = {2019},
isbn = {9781450361729},
publisher = {Association for Computing Machinery},
address = {New York, NY, USA},
url = {https://doi.org/10.1145/3331184.3331265},
doi = {10.1145/3331184.3331265},
booktitle = {Proceedings of the 42nd International ACM SIGIR Conference on Research and Development in Information Retrieval},
pages = {475–484},
numpages = {10},
location = {Paris, France},
series = {SIGIR'19}
}

@inproceedings{aliannejadi2020convai3,
  author = {Aliannejadi, Mohammad and Azzopardi, Leif and Zamani, Hamed and Kanoulas, Evangelos and Thomas, Paul and Craswell, Nick},
title = {Analysing Mixed Initiatives and Search Strategies during Conversational Search},
year = {2021},
isbn = {9781450384469},
publisher = {Association for Computing Machinery},
address = {New York, NY, USA},
url = {https://doi.org/10.1145/3459637.3482231},
doi = {10.1145/3459637.3482231},
booktitle = {Proceedings of the 30th ACM International Conference on Information \& Knowledge Management},
pages = {16–26},
numpages = {11},
location = {Virtual Event, Queensland, Australia},
series = {CIKM '21}
}

@inproceedings{kumar2020clarq,
  title = "{C}lar{Q}: A large-scale and diverse dataset for Clarification Question Generation",
    author = "Kumar, Vaibhav  and
      Black, Alan W",
    editor = "Jurafsky, Dan  and
      Chai, Joyce  and
      Schluter, Natalie  and
      Tetreault, Joel",
    booktitle = "Proceedings of the 58th Annual Meeting of the Association for Computational Linguistics",
    month = jul,
    year = "2020",
    address = "Online",
    publisher = "Association for Computational Linguistics",
    url = "https://aclanthology.org/2020.acl-main.651/",
    doi = "10.18653/v1/2020.acl-main.651",
    pages = "7296--7301"
}

@inproceedings{min2020ambigqa,
  title = "{A}mbig{QA}: Answering Ambiguous Open-domain Questions",
    author = "Min, Sewon  and
      Michael, Julian  and
      Hajishirzi, Hannaneh  and
      Zettlemoyer, Luke",
    editor = "Webber, Bonnie  and
      Cohn, Trevor  and
      He, Yulan  and
      Liu, Yang",
    booktitle = "Proceedings of the 2020 Conference on Empirical Methods in Natural Language Processing (EMNLP)",
    month = nov,
    year = "2020",
    address = "Online",
    publisher = "Association for Computational Linguistics",
    url = "https://aclanthology.org/2020.emnlp-main.466/",
    doi = "10.18653/v1/2020.emnlp-main.466",
    pages = "5783--5797",
}

@inproceedings{kuhn2023clam,
      title={CLAM: Selective Clarification for Ambiguous Questions with Generative Language Models}, 
      author={Lorenz Kuhn and Yarin Gal and Sebastian Farquhar},
      year={2023},
      eprint={2212.07769},
      archivePrefix={arXiv},
      primaryClass={cs.CL},
      url={https://arxiv.org/abs/2212.07769}, 
}

@inproceedings{kim2024apa,
title = "Aligning Language Models to Explicitly Handle Ambiguity",
    author = "Kim, Hyuhng Joon  and
      Kim, Youna  and
      Park, Cheonbok  and
      Kim, Junyeob  and
      Park, Choonghyun  and
      Yoo, Kang Min  and
      Lee, Sang-goo  and
      Kim, Taeuk",
    editor = "Al-Onaizan, Yaser  and
      Bansal, Mohit  and
      Chen, Yun-Nung",
    booktitle = "Proceedings of the 2024 Conference on Empirical Methods in Natural Language Processing",
    month = nov,
    year = "2024",
    address = "Miami, Florida, USA",
    publisher = "Association for Computational Linguistics",
    url = "https://aclanthology.org/2024.emnlp-main.119/",
    doi = "10.18653/v1/2024.emnlp-main.119",
    pages = "1989--2007",
}

@inproceedings{zhang2025futureturn,
  title={Modeling Future Conversation Turns to Teach LLMs to Ask Clarifying Questions},
  author={Michael J.Q. Zhang and W. Bradley Knox and Eunsol Choi},
  journal={ArXiv},
  year={2024},
  volume={abs/2410.13788},
  url={https://api.semanticscholar.org/CorpusID:273403502}
}

@inproceedings{li2024contradoc,
  title = "{C}ontra{D}oc: Understanding Self-Contradictions in Documents with Large Language Models",
    author = "Li, Jierui  and
      Raheja, Vipul  and
      Kumar, Dhruv",
    editor = "Duh, Kevin  and
      Gomez, Helena  and
      Bethard, Steven",
    booktitle = "Proceedings of the 2024 Conference of the North American Chapter of the Association for Computational Linguistics: Human Language Technologies (Volume 1: Long Papers)",
    month = jun,
    year = "2024",
    address = "Mexico City, Mexico",
    publisher = "Association for Computational Linguistics",
    url = "https://aclanthology.org/2024.naacl-long.362/",
    doi = "10.18653/v1/2024.naacl-long.362",
    pages = "6509--6523",
}

@inproceedings{saeidi2018sharc,
  title={Interpretation of Natural Language Rules in Conversational Machine Reading}, 
      author={Marzieh Saeidi and Max Bartolo and Patrick Lewis and Sameer Singh and Tim Rocktäschel and Mike Sheldon and Guillaume Bouchard and Sebastian Riedel},
      year={2018},
      eprint={1809.01494},
      archivePrefix={arXiv},
      primaryClass={cs.CL},
      url={https://arxiv.org/abs/1809.01494}, 
}

@inproceedings{deshpande2025multichallenge,
  title = "{M}ulti{C}hallenge: A Realistic Multi-Turn Conversation Evaluation Benchmark Challenging to Frontier {LLM}s",
    author = "Deshpande, Kaustubh  and
      Sirdeshmukh, Ved  and
      Mols, Johannes Baptist  and
      Jin, Lifeng  and
      Hernandez-Cardona, Ed-Yeremai  and
      Lee, Dean  and
      Kritz, Jeremy  and
      Primack, Willow E.  and
      Yue, Summer  and
      Xing, Chen",
    editor = "Che, Wanxiang  and
      Nabende, Joyce  and
      Shutova, Ekaterina  and
      Pilehvar, Mohammad Taher",
    booktitle = "Findings of the Association for Computational Linguistics: ACL 2025",
    month = jul,
    year = "2025",
    address = "Vienna, Austria",
    publisher = "Association for Computational Linguistics",
    url = "https://aclanthology.org/2025.findings-acl.958/",
    doi = "10.18653/v1/2025.findings-acl.958",
    pages = "18632--18702",
    ISBN = "979-8-89176-256-5",
}

@inproceedings{yao2025taubench,
  title={$\tau$-bench: A Benchmark for Tool-Agent-User Interaction in Real-World Domains}, 
      author={Shunyu Yao and Noah Shinn and Pedram Razavi and Karthik Narasimhan},
      year={2024},
      eprint={2406.12045},
      archivePrefix={arXiv},
      primaryClass={cs.AI},
      url={https://arxiv.org/abs/2406.12045},
}

@inproceedings{xiang2025rmtbench,
  title = "{RMTB}ench: Benchmarking {LLM}s Through Multi-Turn User-Centric Role-Playing",
    author = "Xiang, Hao  and
      Tang, Tianyi  and
      Su, Yang  and
      Yu, Bowen  and
      Yang, An  and
      Huang, Fei  and
      Zhang, Yichang  and
      Lu, Yaojie  and
      Lin, Hongyu  and
      Han, Xianpei  and
      Zhou, Jingren  and
      Lin, Junyang  and
      Sun, Le",
    editor = "Christodoulopoulos, Christos  and
      Chakraborty, Tanmoy  and
      Rose, Carolyn  and
      Peng, Violet",
    booktitle = "Findings of the Association for Computational Linguistics: EMNLP 2025",
    month = nov,
    year = "2025",
    address = "Suzhou, China",
    publisher = "Association for Computational Linguistics",
    url = "https://aclanthology.org/2025.findings-emnlp.730/",
    doi = "10.18653/v1/2025.findings-emnlp.730",
    pages = "13555--13571",
    ISBN = "979-8-89176-335-7",
}

@inproceedings{laban2026lost,
  title={LLMs Get Lost In Multi-Turn Conversation}, 
      author={Philippe Laban and Hiroaki Hayashi and Yingbo Zhou and Jennifer Neville},
      year={2025},
      eprint={2505.06120},
      archivePrefix={arXiv},
      primaryClass={cs.CL},
      url={https://arxiv.org/abs/2505.06120}, 
}

@inproceedings{zheng2023judging,
  author = {Zheng, Lianmin and Chiang, Wei-Lin and Sheng, Ying and Zhuang, Siyuan and Wu, Zhanghao and Zhuang, Yonghao and Lin, Zi and Li, Zhuohan and Li, Dacheng and Xing, Eric P. and Zhang, Hao and Gonzalez, Joseph E. and Stoica, Ion},
title = {Judging LLM-as-a-judge with MT-bench and Chatbot Arena},
year = {2023},
publisher = {Curran Associates Inc.},
address = {Red Hook, NY, USA},
booktitle = {Proceedings of the 37th International Conference on Neural Information Processing Systems},
articleno = {2020},
numpages = {29},
location = {New Orleans, LA, USA},
series = {NIPS '23}
}

@inproceedings{chiang2024chatbot,
  author = {Chiang, Wei-Lin and Zheng, Lianmin and Sheng, Ying and Angelopoulos, Anastasios N. and Li, Tianle and Li, Dacheng and Zhu, Banghua and Zhang, Hao and Jordan, Michael I. and Gonzalez, Joseph E. and Stoica, Ion},
title = {Chatbot arena: an open platform for evaluating LLMs by human preference},
year = {2024},
publisher = {JMLR.org},
booktitle = {Proceedings of the 41st International Conference on Machine Learning},
articleno = {331},
numpages = {30},
location = {Vienna, Austria},
series = {ICML'24}
}

@inproceedings{li2024arenahard,
  author = {Li, Tianle and Chiang, Wei-Lin and Frick, Evan and Dunlap, Lisa and Wu, Tianhao and Zhu, Banghua and Gonzalez, Joseph E. and Stoica, Ion},
title = {From crowdsourced data to high-quality benchmarks: arena-hard and benchbuilder pipeline},
year = {2025},
publisher = {JMLR.org},
booktitle = {Proceedings of the 42nd International Conference on Machine Learning},
articleno = {1340},
numpages = {23},
location = {Vancouver, Canada},
series = {ICML'25}
}

@article{dubois2024alpacaeval,
  title={Length-Controlled AlpacaEval: A Simple Way to Debias Automatic Evaluators}, 
      author={Yann Dubois and Balázs Galambosi and Percy Liang and Tatsunori B. Hashimoto},
      year={2025},
      eprint={2404.04475},
      archivePrefix={arXiv},
      primaryClass={cs.LG},
      url={https://arxiv.org/abs/2404.04475}, 
}

@inproceedings{kwan2024mteval,
  title = "{MT}-Eval: A Multi-Turn Capabilities Evaluation Benchmark for Large Language Models",
    author = "Kwan, Wai-Chung  and
      Zeng, Xingshan  and
      Jiang, Yuxin  and
      Wang, Yufei  and
      Li, Liangyou  and
      Shang, Lifeng  and
      Jiang, Xin  and
      Liu, Qun  and
      Wong, Kam-Fai",
    editor = "Al-Onaizan, Yaser  and
      Bansal, Mohit  and
      Chen, Yun-Nung",
    booktitle = "Proceedings of the 2024 Conference on Empirical Methods in Natural Language Processing",
    month = nov,
    year = "2024",
    address = "Miami, Florida, USA",
    publisher = "Association for Computational Linguistics",
    url = "https://aclanthology.org/2024.emnlp-main.1124/",
    doi = "10.18653/v1/2024.emnlp-main.1124",
    pages = "20153--20177",
}

@inproceedings{liu2023agentbench,
  title={AgentBench: Evaluating LLMs as Agents}, 
      author={Xiao Liu and Hao Yu and Hanchen Zhang and Yifan Xu and Xuanyu Lei and Hanyu Lai and Yu Gu and Hangliang Ding and Kaiwen Men and Kejuan Yang and Shudan Zhang and Xiang Deng and Aohan Zeng and Zhengxiao Du and Chenhui Zhang and Sheng Shen and Tianjun Zhang and Yu Su and Huan Sun and Minlie Huang and Yuxiao Dong and Jie Tang},
      year={2025},
      eprint={2308.03688},
      archivePrefix={arXiv},
      primaryClass={cs.AI},
      url={https://arxiv.org/abs/2308.03688}, 
}

@inproceedings{zhou2024webarena,
  title={WebArena: A Realistic Web Environment for Building Autonomous Agents},
  author={Zhou, Shuyan and Xu, Frank F and Zhu, Hao and Zhou, Xuhui and Lo, Robert and Sridhar, Abishek and Cheng, Xianyi and Bisk, Yonatan and Fried, Daniel and Alon, Uri and others},
  journal={arXiv preprint arXiv:2307.13854},
  url={https://webarena.dev},
  year={2023}
}

@article{mialon2024gaia,
title={{GAIA}: a benchmark for General {AI} Assistants},
author={Gr{\'e}goire Mialon and Cl{\'e}mentine Fourrier and Thomas Wolf and Yann LeCun and Thomas Scialom},
booktitle={The Twelfth International Conference on Learning Representations},
year={2024},
url={https://openreview.net/forum?id=fibxvahvs3}
}

@inproceedings{jimenez2024swebench,
title={{SWE}-bench: Can Language Models Resolve Real-world Github Issues?},
author={Carlos E Jimenez and John Yang and Alexander Wettig and Shunyu Yao and Kexin Pei and Ofir Press and Karthik R Narasimhan},
booktitle={The Twelfth International Conference on Learning Representations},
year={2024},
url={https://openreview.net/forum?id=VTF8yNQM66}
}

@inproceedings{wang2024mint,
   title={MINT: Evaluating LLMs in Multi-turn Interaction with Tools and Language Feedback}, 
      author={Xingyao Wang and Zihan Wang and Jiateng Liu and Yangyi Chen and Lifan Yuan and Hao Peng and Heng Ji},
      year={2024},
      eprint={2309.10691},
      archivePrefix={arXiv},
      primaryClass={cs.CL},
      url={https://arxiv.org/abs/2309.10691}, 
}

@inproceedings{dou2025simulatorarena,
  title = "{S}imulator{A}rena: Are User Simulators Reliable Proxies for Multi-Turn Evaluation of {AI} Assistants?",
    author = "Dou, Yao  and
      Galley, Michel  and
      Peng, Baolin  and
      Kedzie, Chris  and
      Cai, Weixin  and
      Ritter, Alan  and
      Quirk, Chris  and
      Xu, Wei  and
      Gao, Jianfeng",
    editor = "Christodoulopoulos, Christos  and
      Chakraborty, Tanmoy  and
      Rose, Carolyn  and
      Peng, Violet",
    booktitle = "Proceedings of the 2025 Conference on Empirical Methods in Natural Language Processing",
    month = nov,
    year = "2025",
    address = "Suzhou, China",
    publisher = "Association for Computational Linguistics",
    url = "https://aclanthology.org/2025.emnlp-main.1786/",
    doi = "10.18653/v1/2025.emnlp-main.1786",
    pages = "35212--35290",
    ISBN = "979-8-89176-332-6",
}

@inproceedings{shim2026noncollaborative,
  title={Non-Collaborative User Simulators for Tool Agents},
author={Jeonghoon Shim and Woojung Song and Cheyon Jin and Seungwon Kook and Yohan Jo},
booktitle={The Fourteenth International Conference on Learning Representations},
year={2026},
url={https://openreview.net/forum?id=UAUimofy3W}
}

@article{davidson2023usersim,
title={User Simulation with Large Language Models for Evaluating Task-Oriented Dialogue}, 
      author={Sam Davidson and Salvatore Romeo and Raphael Shu and James Gung and Arshit Gupta and Saab Mansour and Yi Zhang},
      year={2023},
      eprint={2309.13233},
      archivePrefix={arXiv},
      primaryClass={cs.CL},
      url={https://arxiv.org/abs/2309.13233}, 
}

@inproceedings{ni2026usersimsurvey,
 title = "A Survey on {LLM}-based Conversational User Simulation",
    author = "Ni, Bo  and
      Wang, Yu  and
      Wang, Leyao  and
      Kveton, Branislav  and
      Dernoncourt, Franck  and
      Xia, Yu  and
      Chen, Hongjie  and
      Luera, Reuben  and
      Basu, Samyadeep  and
      Mukherjee, Subhojyoti  and
      Mathur, Puneet  and
      Ahmed, Nesreen K.  and
      Wu, Junda  and
      Li, Li  and
      Zhang, Huixin  and
      Zhang, Ruiyi  and
      Yu, Tong  and
      Kim, Sungchul  and
      Gu, Jiuxiang  and
      Tu, Zhengzhong  and
      Siu, Alexa  and
      Wang, Zichao  and
      Yoon, Seunghyun  and
      Lipka, Nedim  and
      Park, Namyong  and
      Lin, Zihao  and
      Bui, Trung  and
      Zhao, Yue  and
      Derr, Tyler  and
      Rossi, Ryan A.",
    editor = "Demberg, Vera  and
      Inui, Kentaro  and
      Marquez, Llu{\'i}s",
    booktitle = "Proceedings of the 19th Conference of the {E}uropean Chapter of the {A}ssociation for {C}omputational {L}inguistics (Volume 1: Long Papers)",
    month = mar,
    year = "2026",
    address = "Rabat, Morocco",
    publisher = "Association for Computational Linguistics",
    url = "https://aclanthology.org/2026.eacl-long.200/",
    doi = "10.18653/v1/2026.eacl-long.200",
    pages = "4266--4301",
    ISBN = "979-8-89176-380-7",
}

@article{argyle2023silicon,
  title={Out of One, Many: Using Language Models to Simulate Human Samples},
  author={Lisa P. Argyle and E. Busby and Nancy Fulda and Joshua R Gubler and Christopher Rytting and David Wingate},
  journal={Political Analysis},
  year={2022},
  volume={31},
  pages={337 - 351},
  url={https://api.semanticscholar.org/CorpusID:252280474}
}

@inproceedings{budzianowski2018multiwoz,
 title = "{M}ulti{WOZ} - A Large-Scale Multi-Domain {W}izard-of-{O}z Dataset for Task-Oriented Dialogue Modelling",
    author = "Budzianowski, Pawe{\l}  and
      Wen, Tsung-Hsien  and
      Tseng, Bo-Hsiang  and
      Casanueva, I{\~n}igo  and
      Ultes, Stefan  and
      Ramadan, Osman  and
      Ga{\v{s}}i{\'c}, Milica",
    editor = "Riloff, Ellen  and
      Chiang, David  and
      Hockenmaier, Julia  and
      Tsujii, Jun{'}ichi",
    booktitle = "Proceedings of the 2018 Conference on Empirical Methods in Natural Language Processing",
    month = oct # "-" # nov,
    year = "2018",
    address = "Brussels, Belgium",
    publisher = "Association for Computational Linguistics",
    url = "https://aclanthology.org/D18-1547/",
    doi = "10.18653/v1/D18-1547",
    pages = "5016--5026",
}

@inproceedings{tian2024scicode,
  title={SciCode: A Research Coding Benchmark Curated by Scientists},
author={Minyang Tian and Luyu Gao and Dylan Zhang and Xinan Chen and Cunwei Fan and Xuefei Guo and Roland Haas and Pan Ji and Kittithat Krongchon and Yao Li and Shengyan Liu and Di Luo and Yutao Ma and HAO TONG and Kha Trinh and Chenyu Tian and Zihan Wang and Bohao Wu and Shengzhu Yin and Minhui Zhu and Kilian Lieret and Yanxin Lu and Genglin Liu and Yufeng Du and Tianhua Tao and Ofir Press and Jamie Callan and Eliu A Huerta and Hao Peng},
booktitle={The Thirty-eight Conference on Neural Information Processing Systems Datasets and Benchmarks Track},
year={2024},
url={https://openreview.net/forum?id=ADLaALtdoG}
}

@article{liu2025mattools,
  title={MatTools: Benchmarking Large Language Models for Materials Science Tools}, 
      author={Siyu Liu and Bo Hu and Beilin Ye and Jiamin Xu and David J. Srolovitz and Tongqi Wen},
      year={2025},
      eprint={2505.10852},
      archivePrefix={arXiv},
      primaryClass={cond-mat.mtrl-sci},
      url={https://arxiv.org/abs/2505.10852}, 
}

@inproceedings{chen2024scienceagentbench,
  title={ScienceAgentBench: Toward Rigorous Assessment of Language Agents for Data-Driven Scientific Discovery}, 
      author={Ziru Chen and Shijie Chen and Yuting Ning and Qianheng Zhang and Boshi Wang and Botao Yu and Yifei Li and Zeyi Liao and Chen Wei and Zitong Lu and Vishal Dey and Mingyi Xue and Frazier N. Baker and Benjamin Burns and Daniel Adu-Ampratwum and Xuhui Huang and Xia Ning and Song Gao and Yu Su and Huan Sun},
      year={2025},
      eprint={2410.05080},
      archivePrefix={arXiv},
      primaryClass={cs.CL},
      url={https://arxiv.org/abs/2410.05080}, 
}

@inproceedings{ma2024sciagent,
  title = "{S}ci{A}gent: Tool-augmented Language Models for Scientific Reasoning",
    author = "Ma, Yubo  and
      Gou, Zhibin  and
      Hao, Junheng  and
      Xu, Ruochen  and
      Wang, Shuohang  and
      Pan, Liangming  and
      Yang, Yujiu  and
      Cao, Yixin  and
      Sun, Aixin",
    editor = "Al-Onaizan, Yaser  and
      Bansal, Mohit  and
      Chen, Yun-Nung",
    booktitle = "Proceedings of the 2024 Conference on Empirical Methods in Natural Language Processing",
    month = nov,
    year = "2024",
    address = "Miami, Florida, USA",
    publisher = "Association for Computational Linguistics",
    url = "https://aclanthology.org/2024.emnlp-main.880/",
    doi = "10.18653/v1/2024.emnlp-main.880",
    pages = "15701--15736",
}

@article{bran2024chemcrow,
 title={ChemCrow: Augmenting large-language models with chemistry tools}, 
      author={Andres M Bran and Sam Cox and Oliver Schilter and Carlo Baldassari and Andrew D White and Philippe Schwaller},
      year={2023},
      eprint={2304.05376},
      archivePrefix={arXiv},
      primaryClass={physics.chem-ph},
      url={https://arxiv.org/abs/2304.05376}, 
}

@article{dong2025nl2foam,
  title={Fine-tuning a large language model for automating computational fluid dynamics simulations},
   volume={15},
   ISSN={2095-0349},
   url={http://dx.doi.org/10.1016/j.taml.2025.100594},
   DOI={10.1016/j.taml.2025.100594},
   number={3},
   journal={Theoretical and Applied Mechanics Letters},
   publisher={Elsevier BV},
   author={Dong, Zhehao and Lu, Zhen and Yang, Yue},
   year={2025},
   month=May, pages={100594} 
}

@article{somasekharan2025cfdllmbench,
 itle={{CFDLLMBench}: A Benchmark Suite for Evaluating Large Language Models in Computational Fluid Dynamics},
  author={Somasekharan, Nithin and Yue, Ling and Cao, Yadi and Li, Weichao and Emami, Patrick and Bhargav, Pochinapeddi Sai and Acharya, Anurag and Xie, Xingyu and Pan, Shaowu},
  journal={Journal of Data-centric Machine Learning Research},
  volume={3},
  number={13},
  pages={1--40},
  year={2026},
  month={2},
  publisher={JMLR.org},
  url={https://openreview.net/forum?id=kTcH1MnkjY}
}

@article{chen2024metaopenfoam,
  title={MetaOpenFOAM: an LLM-based multi-agent framework for CFD}, 
      author={Yuxuan Chen and Xu Zhu and Hua Zhou and Zhuyin Ren},
      year={2024},
      eprint={2407.21320},
      archivePrefix={arXiv},
      primaryClass={cs.AI},
      url={https://arxiv.org/abs/2407.21320}, 
}

@article{mudur2025feabench,
       title={FEABench: Evaluating Language Models on Multiphysics Reasoning Ability}, 
      author={Nayantara Mudur and Hao Cui and Subhashini Venugopalan and Paul Raccuglia and Michael P. Brenner and Peter Norgaard},
      year={2025},
      eprint={2504.06260},
      archivePrefix={arXiv},
      primaryClass={cs.AI},
      url={https://arxiv.org/abs/2504.06260}, 
}

@inproceedings{hou2025autofea,
  title = {AutoFEA: Enhancing AI Copilot by Integrating Finite Element Analysis Using Large Language Models with Graph Neural Networks},
author = {Hou, Shifu and Johnson, Rick and Makhija, Ramandeep and Chen, Lingwei and Ye, Yanfang},
booktitle = {Proceedings of the AAAI Conference on Artificial Intelligence (AAAI)},
pages = {24078--24085},
year = {2025}
}

@article{deotale2026allfem,
      title={ALL-FEM: Agentic Large Language models Fine-tuned for Finite Element Methods}, 
      author={Rushikesh Deotale and Adithya Srinivasan and Yuan Tian and Tianyi Zhang and Pavlos Vlachos and Hector Gomez},
      year={2026},
      eprint={2603.21011},
      archivePrefix={arXiv},
      primaryClass={cs.CE},
      url={https://arxiv.org/abs/2603.21011}, 
}

@inproceedings{zhang2024honeycomb,
  title={HoneyComb: A Flexible LLM-Based Agent System for Materials Science}, 
      author={Huan Zhang and Yu Song and Ziyu Hou and Santiago Miret and Bang Liu},
      year={2024},
      eprint={2409.00135},
      archivePrefix={arXiv},
      primaryClass={cs.CL},
      url={https://arxiv.org/abs/2409.00135}, 
}

@article{ni2024mechagents,
 title={MechAgents: Large language model multi-agent collaborations can solve mechanics problems, generate new data, and integrate knowledge}, 
      author={Bo Ni and Markus J. Buehler},
      year={2023},
      eprint={2311.08166},
      archivePrefix={arXiv},
      primaryClass={cs.AI},
      url={https://arxiv.org/abs/2311.08166}, 
}

@book{munson2020fundamentals,
  title={Munson, Young and Okiishi's Fundamentals of Fluid Mechanics},
  author={Gerhart, A.L. and Hochstein, J.I. and Gerhart, P.M.},
  isbn={9781119597308},
  lccn={2020018704},
  url={https://books.google.com/books?id=F6ALEAAAQBAJ},
  year={2020},
  publisher={Wiley}
}

@book{white2021fluid,
  title={Fluid Mechanics},
  author={White, F.M.},
  isbn={9780071311212},
  lccn={2010481187},
  series={McGraw-Hill series in mechanical engineering},
  url={https://books.google.com/books?id=n43PbwAACAAJ},
  year={2011},
  publisher={McGraw-Hill}
}

@article{cfdllmbench,
  title={{CFDLLMBench}: A Benchmark Suite for Evaluating Large Language Models in Computational Fluid Dynamics},
  author={Somasekharan, Nithin and Yue, Ling and Cao, Yadi and Li, Weichao and Emami, Patrick and Bhargav, Pochinapeddi Sai and Acharya, Anurag and Xie, Xingyu and Pan, Shaowu},
  journal={Journal of Data-centric Machine Learning Research},
  volume={3},
  number={13},
  pages={1--40},
  year={2026},
  month={2},
  publisher={JMLR.org},
  url={https://openreview.net/forum?id=kTcH1MnkjY}
}

@book{logg2012fenics,
author = {Logg, Anders and Wells, Garth and Mardal, Kent-Andre},
year = {2011},
month = {04},
pages = {},
title = {Automated Solution of Differential Equations by the Finite Element Method: The FEniCS Book},
volume = {84},
isbn = {978-3-642-23098-1},
journal = {Lecture Notes in Computational Science and Engineering},
doi = {10.1007/978-3-642-23099-8}
}

@book{beer2021mechanics,
  title={Mechanics of Materials},
  author={Beer, F. and Jr. Johnston, E.R. and DeWolf, J. and Mazurek, D.},
  isbn={9780073380285},
  lccn={2010037852},
  url={https://books.google.com/books?id=g635kQAACAAJ},
  year={2011},
  publisher={McGraw-Hill Education}
}

@book{goodno2018mechanics,
  title={Mechanics of Materials},
  author={Gere, J.M. and Goodno, B.J.},
  isbn={9780534553975},
  lccn={2008923451},
  url={https://books.google.com/books?id=EaYTLOKbI8UC},
  year={2008},
  publisher={Cengage Learning}
}

@book{ugural2021advanced,
  title={Advanced Mechanics of Materials and Applied Elasticity},
  author={Ugural, A.C. and Fenster, S.K.},
  isbn={9780137079810},
  lccn={2011012705},
  series={International Series in the Physical and Chemical Engineering Sciences},
  url={https://books.google.com/books?id=vLqkydSJ8vEC},
  year={2011},
  publisher={Pearson Education}
}

@misc{dhondt1998calculix,
  author       = {Guido Dhondt and Klaus Wittig},
  title        = {{CalculiX}: A Three-Dimensional Structural Finite Element Program},
  year         = {1998},
  url          = {https://www.calculix.de/},
  note         = {Software, accessed 2026-04-12}
}

@book{callister2018materials,
  title={Materials Science and Engineering: An Introduction, 9th Edition: Ninth Edition},
  author={Callister, W.D. and Rethwisch, D.G.},
  isbn={9781118476543},
  lccn={2013051260},
  url={https://books.google.com/books?id=TmxbAgAAQBAJ},
  year={2013},
  publisher={John Wiley and Sons, Incorporated}
}

@book{askeland2025science,
  title={The Science and Engineering of Materials, SI Edition},
  author={Askeland, D.R. and Fulay, P.P. and Wright, W.J.},
  isbn={9780495668022},
  lccn={2010922628},
  url={https://books.google.com/books?id=fJanis-q6GkC},
  year={2011},
  publisher={Cengage Learning}
}

@book{shackelford2021materials,
  title={Introduction to Materials Science for Engineers},
  author={Shackelford, J.F.},
  isbn={9780131276192},
  lccn={2004044322},
  url={https://books.google.com/books?id=1nbDQgAACAAJ},
  year={2005},
  publisher={Pearson/Prentice Hall}
}

@article{zaki2024mascqa,
    author = {Zaki, Mohd and Jayadeva and Mausam and Krishnan, N. M. Anoop},
    title = {MaScQA: investigating materials science knowledge of large language models},
    journal = {Digital Discovery},
    volume = {3},
    number = {2},
    pages = {313-327},
    year = {2024},
    month = {02},
    issn = {2635-098X},
    doi = {10.1039/d3dd00188a},
    url = {https://doi.org/10.1039/d3dd00188a},
    eprint = {https://pubs.rsc.org/dd/article-pdf/3/2/313/9050078/d3dd00188a.pdf},
}

@article{zhang2025matscibench,
 author = {Zhang, Junkai and Gan, Jingru and Wang, Xiaoxuan and Jia, Zian and Gu, Changquan and Chen, Jianpeng and Zhu, Yanqiao and Ma, Mingyu Derek and Zhou, Dawei and Li, Ling and Wang, Wei},
title = {MatSciBench: Benchmarking the Reasoning Ability of Large Language Models in Materials Science},
year = {2026},
isbn = {9798400722592},
publisher = {Association for Computing Machinery},
address = {New York, NY, USA},
url = {https://doi.org/10.1145/3770855.3818888},
doi = {10.1145/3770855.3818888},
booktitle = {Proceedings of the 32nd ACM SIGKDD Conference on Knowledge Discovery and Data Mining V.2},
pages = {12868–12879},
numpages = {12},
location = {Republic of Korea},
series = {KDD '26}
}

@inproceedings{bai2024mtbench101,
 title = "{MT}-Bench-101: A Fine-Grained Benchmark for Evaluating Large Language Models in Multi-Turn Dialogues",
    author = "Bai, Ge  and
      Liu, Jie  and
      Bu, Xingyuan  and
      He, Yancheng  and
      Liu, Jiaheng  and
      Zhou, Zhanhui  and
      Lin, Zhuoran  and
      Su, Wenbo  and
      Ge, Tiezheng  and
      Zheng, Bo  and
      Ouyang, Wanli",
    editor = "Ku, Lun-Wei  and
      Martins, Andre  and
      Srikumar, Vivek",
    booktitle = "Proceedings of the 62nd Annual Meeting of the Association for Computational Linguistics (Volume 1: Long Papers)",
    month = aug,
    year = "2024",
    address = "Bangkok, Thailand",
    publisher = "Association for Computational Linguistics",
    url = "https://aclanthology.org/2024.acl-long.401/",
    doi = "10.18653/v1/2024.acl-long.401",
    pages = "7421--7454",
}

@inproceedings{wang2024scibench,
  author = {Wang, Xiaoxuan and Hu, Ziniu and Lu, Pan and Zhu, Yanqiao and Zhang, Jieyu and Subramaniam, Satyen and Loomba, Arjun R. and Zhang, Shichang and Sun, Yizhou and Wang, Wei},
title = {SCIBENCH: evaluating college-level scientific problem-solving abilities of large language models},
year = {2024},
publisher = {JMLR.org},
booktitle = {Proceedings of the 41st International Conference on Machine Learning},
articleno = {2072},
numpages = {28},
location = {Vienna, Austria},
series = {ICML'24}
}

@inproceedings{sun2024scieval,
  author = {Sun, Liangtai and Han, Yang and Zhao, Zihan and Ma, Da and Shen, Zhennan and Chen, Baocai and Chen, Lu and Yu, Kai},
title = {SciEval: a multi-level large language model evaluation benchmark for scientific research},
year = {2024},
isbn = {978-1-57735-887-9},
publisher = {AAAI Press},
url = {https://doi.org/10.1609/aaai.v38i17.29872},
doi = {10.1609/aaai.v38i17.29872},
booktitle = {Proceedings of the Thirty-Eighth AAAI Conference on Artificial Intelligence and Thirty-Sixth Conference on Innovative Applications of Artificial Intelligence and Fourteenth Symposium on Educational Advances in Artificial Intelligence},
articleno = {2124},
numpages = {9},
series = {AAAI'24/IAAI'24/EAAI'24}
}

@inproceedings{fu2025questbench,
  title={{QuestBench}: Can LLMs ask the right question to acquire information in reasoning tasks?},
  author={Li, Belinda Z and Kim, Been and Wang, Zi},
  booktitle={Advances in Neural Information Processing Systems (NeurIPS)},
  year={2025}
}

@article{gan2024clarqllm,
title={ClarQ-LLM: A Benchmark for Models Clarifying and Requesting Information in Task-Oriented Dialog}, 
      author={Yujian Gan and Changling Li and Jinxia Xie and Luou Wen and Matthew Purver and Massimo Poesio},
      year={2024},
      eprint={2409.06097},
      archivePrefix={arXiv},
      primaryClass={cs.CL},
      url={https://arxiv.org/abs/2409.06097}, 
}

@article{singhal2025profbench,
title={ProfBench: Multi-Domain Rubrics requiring Professional Knowledge to Answer and Judge},
author={Zhilin Wang and Jaehun Jung and Ximing Lu and Shizhe Diao and Ellie Evans and Jiaqi Zeng and Pavlo Molchanov and Yejin Choi and Jan Kautz and Yi Dong},
booktitle={The Fourteenth International Conference on Learning Representations},
year={2026},
url={https://openreview.net/forum?id=VwNzKPqBxk}
}

@inproceedings{hashemi2024llmrubric,
  title = "{LLM}-Rubric: A Multidimensional, Calibrated Approach to Automated Evaluation of Natural Language Texts",
    author = "Hashemi, Helia  and
      Eisner, Jason  and
      Rosset, Corby  and
      Van Durme, Benjamin  and
      Kedzie, Chris",
    editor = "Ku, Lun-Wei  and
      Martins, Andre  and
      Srikumar, Vivek",
    booktitle = "Proceedings of the 62nd Annual Meeting of the Association for Computational Linguistics (Volume 1: Long Papers)",
    month = aug,
    year = "2024",
    address = "Bangkok, Thailand",
    publisher = "Association for Computational Linguistics",
    url = "https://aclanthology.org/2024.acl-long.745/",
    doi = "10.18653/v1/2024.acl-long.745",
    pages = "13806--13834",
}

@article{barres2025tau2,
      title={$\tau^2$-Bench: Evaluating Conversational Agents in a Dual-Control Environment}, 
      author={Victor Barres and Honghua Dong and Soham Ray and Xujie Si and Karthik Narasimhan},
      year={2025},
      eprint={2506.07982},
      archivePrefix={arXiv},
      primaryClass={cs.AI},
      url={https://arxiv.org/abs/2506.07982}, 
}

@article{hathidara2026mirrorbench,
author = {Hathidara, Ashutosh and Yu, Julien and Senthil, Vaishali and Schreiber, Sebastian and Ankisettipalli, Anil Babu},
title = {MirrorBench: A Benchmark to Evaluate Conversational User-Proxy Agents for Human-Likeness},
year = {2026},
isbn = {9798400722592},
publisher = {Association for Computing Machinery},
address = {New York, NY, USA},
url = {https://doi.org/10.1145/3770855.3817496},
doi = {10.1145/3770855.3817496},
booktitle = {Proceedings of the 32nd ACM SIGKDD Conference on Knowledge Discovery and Data Mining V.2},
pages = {8978–8988},
numpages = {11},
location = {Republic of Korea},
series = {KDD '26}
}

@article{luo2024reliableusersim,
  title = "Reliable {LLM}-based User Simulator for Task-Oriented Dialogue Systems",
    author = "Sekulic, Ivan  and
      Terragni, Silvia  and
      Guimar{\~a}es, Victor  and
      Khau, Nghia  and
      Guedes, Bruna  and
      Filipavicius, Modestas  and
      Manso, Andre Ferreira  and
      Mathis, Roland",
    editor = "Graham, Yvette  and
      Liu, Qun  and
      Lampouras, Gerasimos  and
      Iacobacci, Ignacio  and
      Madden, Sinead  and
      Khalid, Haider  and
      Qureshi, Rameez",
    booktitle = "Proceedings of the 1st Workshop on Simulating Conversational Intelligence in Chat (SCI-CHAT 2024)",
    month = mar,
    year = "2024",
    address = "St. Julians, Malta",
    publisher = "Association for Computational Linguistics",
    url = "https://aclanthology.org/2024.scichat-1.3/",
    doi = "10.18653/v1/2024.scichat-1.3",
    pages = "19--35",
}

@misc{ashton2025fluidintelligenceforwardlook,
      title={Fluid Intelligence: A Forward Look on AI Foundation Models in Computational Fluid Dynamics}, 
      author={Neil Ashton and Johannes Brandstetter and Siddhartha Mishra},
      year={2025},
      eprint={2511.20455},
      archivePrefix={arXiv},
      primaryClass={physics.flu-dyn},
      url={https://arxiv.org/abs/2511.20455}, 
}

@inproceedings{tyen-etal-2024-llms,
    title = "{LLM}s cannot find reasoning errors, but can correct them given the error location",
    author = "Tyen, Gladys  and
      Mansoor, Hassan  and
      Carbune, Victor  and
      Chen, Peter  and
      Mak, Tony",
    editor = "Ku, Lun-Wei  and
      Martins, Andre  and
      Srikumar, Vivek",
    booktitle = "Findings of the Association for Computational Linguistics: ACL 2024",
    month = aug,
    year = "2024",
    address = "Bangkok, Thailand",
    publisher = "Association for Computational Linguistics",
    url = "https://aclanthology.org/2024.findings-acl.826/",
    doi = "10.18653/v1/2024.findings-acl.826",
    pages = "13894--13908",
}

@misc{anthropic2026claude46,
  title        = {Claude Sonnet 4.6 System Card},
  author       = {{Anthropic}},
  year         = {2026},
  month        = feb,
  howpublished = {\url{https://www.anthropic.com/claude-sonnet-4-6-system-card}},
  note         = {System card, February 17, 2026.}
}

@article{google2025gemini25,
  title   = {Gemini 2.5: Pushing the Frontier with Advanced Reasoning, Multimodality, Long Context, and Next Generation Agentic Capabilities},
  author  = {{Gemini Team, Google DeepMind}},
eprint={2507.06261},
      archivePrefix={arXiv},
      primaryClass={cs.CL},
      url={https://arxiv.org/abs/2507.06261}, 
}

@misc{openai2025gpt52,
  title        = {Update to GPT-5 System Card: GPT-5.2},
  author       = {{OpenAI}},
  year         = {2025},
  month        = dec,
  howpublished = {\url{https://openai.com/index/gpt-5-system-card-update-gpt-5-2/}},
  note         = {System card update, December 11, 2025.}
}

@article{openai2025gptoss,
  title   = {gpt-oss-120b \& gpt-oss-20b Model Card},
  author  = {{OpenAI}},
year={2025},
      eprint={2508.10925},
      archivePrefix={arXiv},
      primaryClass={cs.CL},
      url={https://arxiv.org/abs/2508.10925}, 
}

@article{pandey2025openfoamgpt,
  author = {Pandey, Sandeep and Xu, Ran and Wang, Wenkang and Chu, Xu},
    title = {OpenFOAMGPT: A retrieval-augmented large language model (LLM) agent for OpenFOAM-based computational fluid dynamics},
    journal = {Physics of Fluids},
    volume = {37},
    number = {3},
    pages = {035120},
    year = {2025},
    month = {03},
    issn = {1070-6631},
    doi = {10.1063/5.0257555},
    url = {https://doi.org/10.1063/5.0257555},
    eprint = {https://pubs.aip.org/aip/pof/article-pdf/doi/10.1063/5.0257555/20423647/035120_1_5.0257555.pdf},
}

@inproceedings{yue2025foam,
  title={Foam-Agent: A Multi-Agent Framework for Automating OpenFOAM-based CFD Simulation},
  author={Yue, Ling and Somasekharan, Nithin and Cao, Yadi and Pan, Shaowu},
  booktitle={NeurIPS 2025 Workshop ML4PS},
  year={2025}
}

@article{somasekharan2026cfdllmbench,
title={{CFDLLMBench}: A Benchmark Suite for Evaluating Large Language Models in Computational Fluid Dynamics},
  author={Somasekharan, Nithin and Yue, Ling and Cao, Yadi and Li, Weichao and Emami, Patrick and Bhargav, Pochinapeddi Sai and Acharya, Anurag and Xie, Xingyu and Pan, Shaowu},
  journal={Journal of Data-centric Machine Learning Research},
  volume={3},
  number={13},
  pages={1--40},
  year={2026},
  month={2},
  publisher={JMLR.org},
  url={https://openreview.net/forum?id=kTcH1MnkjY}
}
}

\appendix
\appendix
\section{Limitations}
\label{app:limitations}
\benchmark{} is limited to four computational-science domains and to English-language, text-only prompts at undergraduate-to-early-graduate difficulty; the absolute numbers therefore should not be extrapolated to other domains, modalities, or research-level tasks. The dataset contains roughly 1{,}000 cases, reflecting the fact that scientific task-formulation data are sparse and substantially harder to construct than standard NLP corpora. We have not yet conducted a human clarification study.

\section{Broader impacts}
\label{app:broader-impacts}
\benchmark{} may have positive societal effects by helping researchers and developers identify silent assumptions in scientific AI workflows before they lead to difficult-to-audit or irreproducible computational results. By measuring whether models ask clarifying questions before finalizing a task specification, the benchmark is intended to support more reliable human-AI interaction in scientific settings. The main negative impact is the possible misuse or overinterpretation of benchmark scores. High performance on \benchmark{} could be taken as evidence that an assistant is ready for autonomous scientific deployment, even though the benchmark evaluates a specific interaction protocol and does not certify scientific correctness, safety, or downstream decision quality. In intended use, incorrect model outputs could still lead users to trust poorly specified simulations, irreproducible workflows, or misleading scientific recommendations. There is also a risk that systems could be optimized for simulated clarification behavior rather than for robust collaboration with human experts. We reduce these risks by making the dataset, rubric, and evaluation code inspectable, by reporting conversation-grounded resolution separately from final correctness, and by describing the benchmark scope explicitly. The released benchmark does not contain personal data, sensitive human-subject data, or dual-use experimental protocols, and we do not identify a direct path to privacy, security, surveillance, or disinformation harms beyond the general risks of deploying scientific AI assistants without appropriate expert oversight.
\section{Prompt Templates}
\label{app:prompt-templates}

This appendix provides the exact prompt templates used in the conversational evaluation framework.

The guided and unguided assistant system prompts are reproduced verbatim
in \cref{app:guided-vs-unguided}, alongside the side-by-side comparison
of the two conditions on \textsc{Gemini 2.5 Pro}.

\subsection{Simulated User Prompt}

\begin{promptbox}{Simulated user system prompt}
You are a User who knows the COMPLETE requirement for a task.

The Agent only has an INCOMPLETE version and is asking clarification questions.
You have access to the COMPLETE requirement and should answer based strictly on that.

IMPORTANT GUIDELINES:
1. Answer questions based on the COMPLETE user requirement.
2. Be specific and provide concrete values if they exist in the COMPLETE requirement.
3. If the complete requirement doesn't specify something, tell the agent: "Make a reasonable assumption based on best practices."
4. Be consistent with previous clarifications.

CONTEXT:
<FULL_CONTEXT>

CRITICAL INSTRUCTIONS:
- Be concise.
- Do not explain your reasoning.
- Do not provide information not asked for.
- Do not make up new constraints that aren't in the complete requirement.
- Answer like a human user would.
\end{promptbox}

\begin{promptbox}{Simulated user query prompt}
The Agent asks:
"<QUESTION>"

Please provide a clear answer based on the COMPLETE requirement.
\end{promptbox}

Here, \texttt{<FULL\_CONTEXT>} is instantiated with the incomplete request, the hidden complete requirement, and the prior clarification history in the following form:

\begin{promptbox}{Simulator context template}
<incomplete_requirement>
<INCOMPLETE_REQUIREMENT>
</incomplete_requirement>

<complete_requirement>
<COMPLETE_REQUIREMENT>
</complete_requirement>

<clarifications>
<exchange>
<question>...</question>
<answer>...</answer>
</exchange>
...
</clarifications>
\end{promptbox}

\subsection{Forced Finalization Prompt}
\label{app:forced_finilization}

When a conversation reaches the turn cap without an explicit finalization, we append the following instruction to the assistant system prompt to force a final specification:

\begin{promptbox}{Forced-finalization suffix}
CRITICAL: Time is up. Do not ask any more questions. Output the final specification now based on current information. Do not include the '[COMPLETE]' tag.
\end{promptbox}

\section{Capability, Robustness, and Usability.}
\label{sec:pareto_axis_define}
CGRR gives the primary success criterion, but it does not explain why a model succeeds or fails. We therefore report three diagnostic axes, used in the Pareto analysis. \textbf{Capability} measures whether the model elicits the right information and produces a complete specification. Let $q_{ij}$ indicate whether issue $j$ is explicitly surfaced in the conversation, $Q_i$ be the number of clarification questions, $Q_i^{\mathrm{rel}}$ the number targeting annotated issues, and $p_{ij}$ indicate whether required specification field $j$ is correctly instantiated. We define
\begin{equation}
\mathrm{CR}_i = \frac{1}{m_i}\sum_{j=1}^{m_i} q_{ij}, \qquad
\mathrm{CP}_i = \frac{Q_i^{\mathrm{rel}}}{\max(Q_i,1)}, \qquad
\mathrm{PC}_i = \frac{1}{m_i}\sum_{j=1}^{m_i} p_{ij}.
\label{eq:capability}
\end{equation}
Here CR is Clarification Recall, CP is Clarification Precision, and PC is Plan Completeness. For inconsistency cases, we also report Detection Recall (DR), defined as CR restricted to planted conflicts.

\textbf{Robustness} measures whether the model avoids silent or inconsistent behavior. The \emph{Assumption Rate} is
\begin{equation}
\mathrm{AR}_i = \frac{1}{m_i}\sum_{j=1}^{m_i} \mathbf{1}\!\left[r_{ij}=1 \wedge g_{ij}=0\right],
\label{eq:assumption-rate}
\end{equation}
where lower is better. We also report Error Detection Rate (ED), the fraction of planted issues explicitly flagged before finalization, and Memory Consistency Rate (MCR), which is one when the final specification does not contradict information established during dialogue. Usability is measured by Intent Capture Rate (ICR), i.e., $\iota_i$, which separates intent drift from resolution and grounding failures.

For compact visualization, we aggregate these diagnostics into axis scores:
\begin{equation}
\mathrm{Cap}_i = \tfrac{1}{3}\!\left(\mathrm{CR}_i + \mathrm{CP}_i + \mathrm{PC}_i\right), \qquad
\mathrm{Rob}_i = \tfrac{1}{3}\!\left[(1-\mathrm{AR}_i) + \mathrm{ED}_i + \mathrm{MCR}_i\right], \qquad
\mathrm{Use}_i = \mathrm{ICR}_i.
\label{eq:pareto-axes}
\end{equation}
Benchmark level axis scores are macro averages over cases. 
\section{Additional Results}
\label{app:additional-results}

\subsection{General numeric questions versus tool-use prompts}
\label{app:general-vs-tooluse}

The main benchmark results in \cref{fig:main-outcomes} aggregate each domain over both textbook-style numeric prompts and tool-oriented prompts. These two prompt types exercise different skills: numeric/PDE reasoning versus scientific software invocation. Therefore, we split the metric across these two groups: The \emph{general numeric} group contains textbook-style problems that do not assume a specific simulation stack and the \emph{tool-use} group contains prompts that presuppose a concrete simulation tool or solver setup. We place \textsc{pde} in the tool-use group because the PDE component is solver-oriented. CGRR and SRR are computed exactly as in the main paper.

\begin{figure*}[h]
    \centering
    \includegraphics[width=\textwidth]{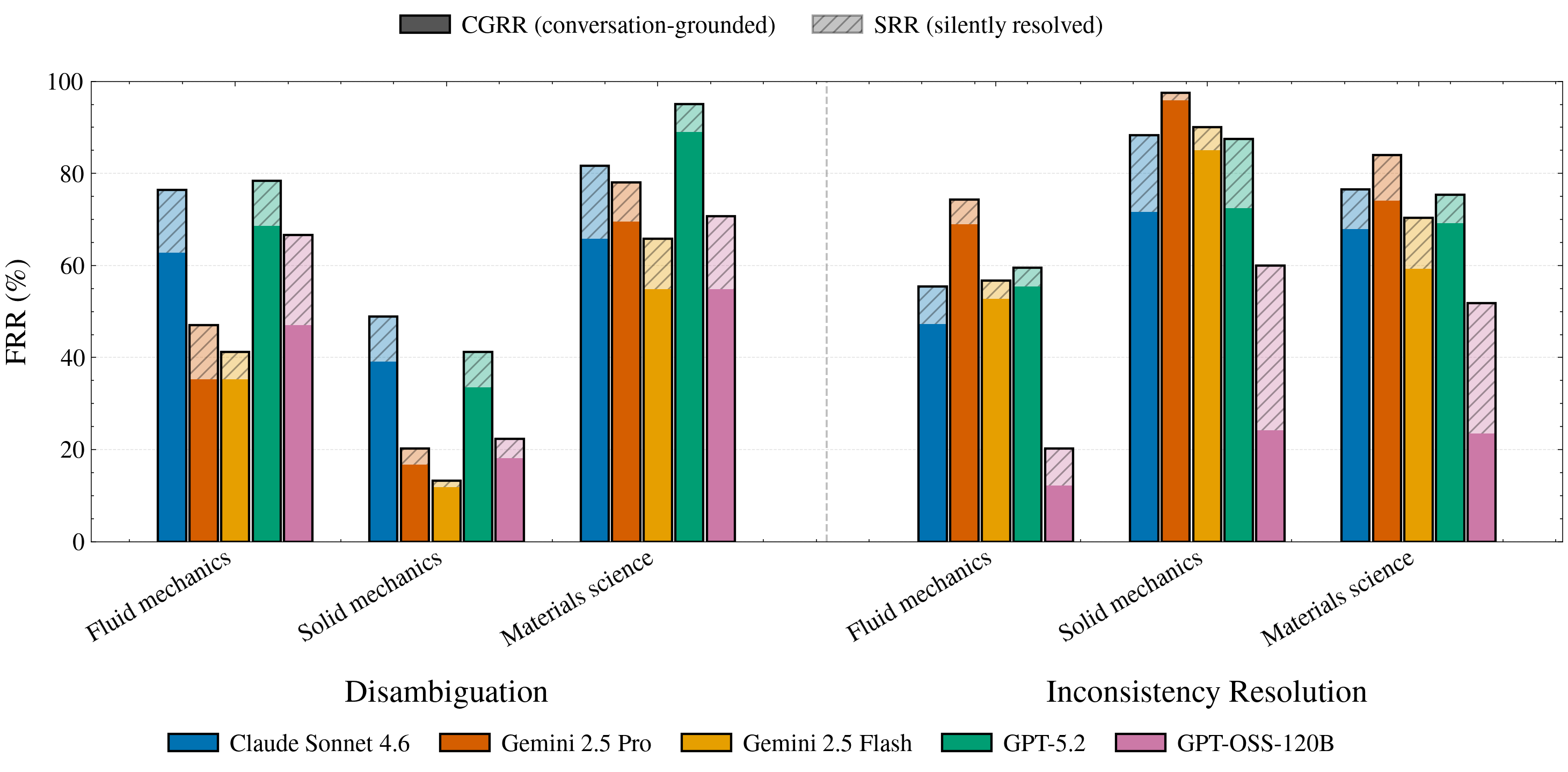}
    \caption{Outcomes on \emph{general numeric} prompts (textbook-style problems without a fixed tool stack). Each bar decomposes outcomes into Conversation-Grounded Resolution Rate (CGRR, colored), Silent Resolution Rate (SRR, grey), and unresolved cases; the bar top is the Final Resolution Rate (FRR). Three domains are available in this split (fluid mechanics, solid mechanics, materials science).}
    \label{fig:appendix-general-numeric}
\end{figure*}

\begin{figure*}[h]
    \centering
    \includegraphics[width=\textwidth]{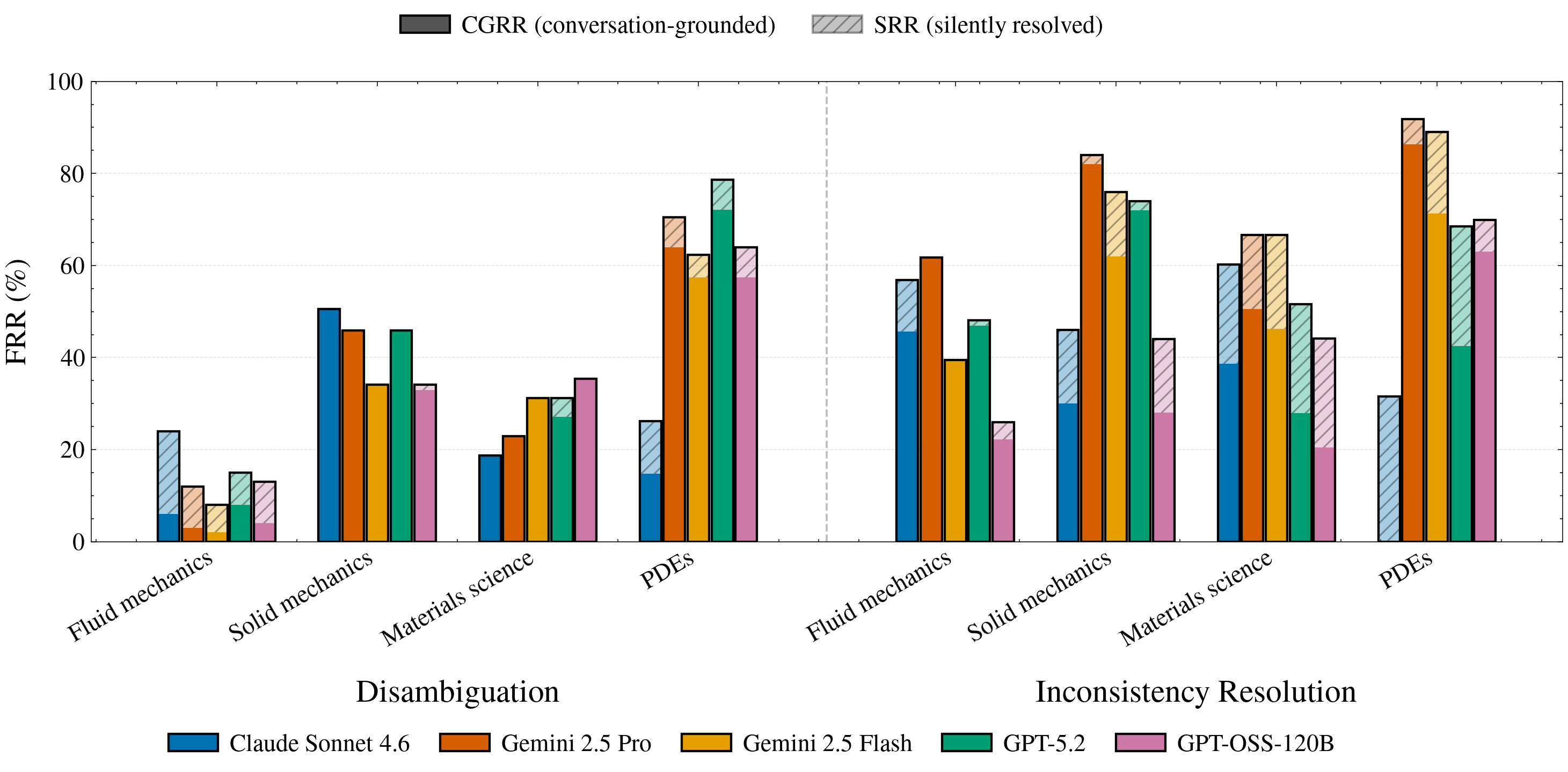}
    \caption{Outcomes on \emph{tool-use} prompts (OpenFOAM, FEA, materials-science tools, and PDE solver setup). Bars use the same decomposition as \cref{fig:appendix-general-numeric}. PDEs are grouped here because the PDE component is solver-oriented and has no textbook-style counterpart.}
    \label{fig:appendix-tooluse}
\end{figure*}

Two qualitative patterns are worth noting. First, the headline failure mode: a large FRR-CGRR gap in disambiguation persists in both splits. Silent resolution is not an artifact of either prompt style alone, but is visible whenever information is missing regardless of whether the downstream task is a textbook calculation or a tool invocation. Second, the splits differ in where Capability pressure concentrates. Tool-use prompts carry additional hidden state (solver choice, mesh/discretization conventions, units and flags expected by the tool), which tends to produce more opportunities for silent assumption and correspondingly wider FRR-CGRR gaps in the disambiguation row; conversely, inconsistency resolution remains comparatively easy across both splits because planted conflicts are locally visible in the prompt. Taken together, these per-prompt-type results reinforce the main-paper conclusion---conversation-grounded success and end-state success come apart---and show that the gap is not driven by any single prompt style.

\subsection{Per-domain breakdown}
\label{app:per-domain}

\Cref{fig:ontology-outcomes} in the main paper reports performance broken
down by ontology component, with the denominator restricted to issues whose
component label matches the bar. For completeness, we also report the same
outcome metrics aggregated by scientific domain rather than by ontology
component.

\paragraph{Per-domain metric definitions.}
Let $\mathcal{I}=\{(i,j)\mid i\in\mathcal{D},\,1\le j\le m_i\}$ index every
(case, planted-issue) pair in the benchmark, with $r_{ij}, g_{ij}, \iota_i$
defined as in \cref{sec:metrics}, and let $d_i$ be the scientific domain of
case $i$ (one of fluid mechanics, solid mechanics, materials science, or
PDEs). The per-domain rates restricted to domain $d$ are
\begin{align*}
\mathrm{FRR}(d)
  &= \frac{\sum_{(i,j)\in\mathcal{I}} \mathbf{1}[d_i=d \,\wedge\, \iota_i=1 \,\wedge\, r_{ij}=1]}
          {\sum_{(i,j)\in\mathcal{I}} \mathbf{1}[d_i=d]}, \\
\mathrm{CGRR}(d)
  &= \frac{\sum_{(i,j)\in\mathcal{I}} \mathbf{1}[d_i=d \,\wedge\, \iota_i=1 \,\wedge\, r_{ij}=1 \,\wedge\, g_{ij}=1]}
          {\sum_{(i,j)\in\mathcal{I}} \mathbf{1}[d_i=d]}, \\
\mathrm{SRR}(d) &= \mathrm{FRR}(d) - \mathrm{CGRR}(d).
\end{align*}
The denominator counts individual planted issues rather than cases: one
missing entity for disambiguation, one planted contradiction for
inconsistency resolution. A case with $m_i$ planted issues contributes $m_i$
observations rather than one. Intent-capture gating ($\iota_i$) is applied
at the case-level, exactly as in \cref{sec:metrics}. Each issue enters
exactly one domain.

\begin{figure}[h]
\centering
\includegraphics[width=\linewidth]{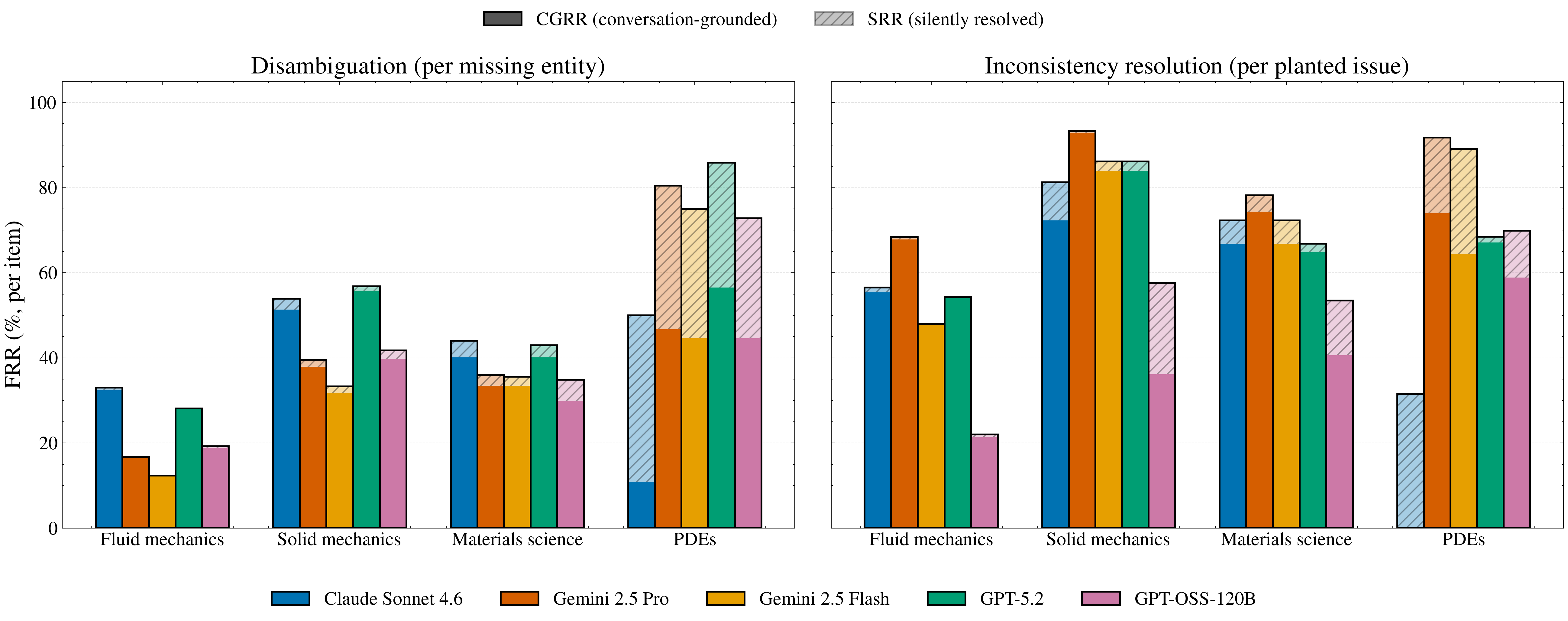}
\caption{Per-domain breakdown ($\mathrm{FRR}(d)$ and $\mathrm{CGRR}(d)$).
Denominator: total missing entities or planted inconsistencies per
(domain, model, task).}
\label{fig:appendix-domain-peritem}
\end{figure}

\paragraph{Per-domain aggregation.}
\Cref{fig:appendix-domain-peritem} reports the per-domain breakdown.
Domains with high issue-counts per case (solid mechanics, fluid mechanics)
compress visually relative to \cref{fig:main-outcomes}; domains with few
issues per case (PDEs) match the case-level numbers closely. The
qualitative ordering across models and the FRR--CGRR gap pattern are
preserved.

\subsection{Full domain-level results}
\label{app:full-domain-results}

\Cref{tab:appendix_disambiguation,tab:appendix_inconsistency} report the full per-domain breakdown of all outcome and diagnostic metrics used in the paper. \Cref{tab:appendix_disambiguation} covers \emph{disambiguation} cases, where planted ambiguities (missing geometry, boundary/initial conditions, material properties, numerical controls, or target outputs) must be elicited through conversation before a valid specification can be produced. \Cref{tab:appendix_inconsistency} covers \emph{inconsistency resolution} cases, where the initial user request contains planted conflicts (e.g., a boundary condition that contradicts the stated geometry or a governing model that contradicts the stated property data) that must be explicitly detected and resolved before finalization. Both tables cover all five guided-mode models (\textsc{Claude Sonnet~4.6}, \textsc{Gemini~2.5~Pro}, \textsc{Gemini~2.5~Flash}, \textsc{GPT-5.2}, \textsc{GPT-OSS-120B}) and report every metric defined in the metrics section: FRR, CGRR, SRR, Clarification Recall (CR), Clarification Precision (CP), Plan Completeness (PC), Assumption Rate (AR, lower is better), Intent Capture Rate (ICR), and Memory Consistency Rate (MCR). The inconsistency-resolution table additionally reports Error Detection Rate (ED), the fraction of planted conflicts that the assistant explicitly flags in dialogue before finalization.


\begin{table}[htbp]
\centering
\footnotesize
\setlength{\tabcolsep}{3pt}
\caption{Full domain-level results on \emph{disambiguation} (all values in \%). Columns follow the metrics section: FRR (Final Resolution Rate), CGRR (Conversation-Grounded Resolution Rate), SRR (Silent Resolution Rate), CR (Clarification Recall), CP (Clarification Precision), PC (Plan Completeness), AR (Assumption Rate, $\downarrow$ is better; reported as a case-level $(1-\mathrm{CR})\cdot\mathrm{FR}$ estimate because the CSV is at case granularity), ICR (Intent Capture Rate), MCR (Memory Consistency Rate). \textbf{Bold} = best per (domain, metric); SRR and AR use the minimum.}
\label{tab:appendix_disambiguation}
\resizebox{\textwidth}{!}{%
\begin{tabular}{llrrrrrrrrr}
\toprule
\textbf{Domain} & \textbf{Model} & \textbf{FRR} & \textbf{CGRR} & \textbf{SRR} & \textbf{CR} & \textbf{CP} & \textbf{PC} & \textbf{AR$\downarrow$} & \textbf{ICR} & \textbf{MCR} \\
\midrule
\multirow{5}{*}{\textit{Fluid mechanics}} & Claude Sonnet 4.6 & \textbf{49.7} & 25.2 & 16.6 & 52.2 & \textbf{88.4} & \textbf{79.4} & 17.7 & \textbf{47.7} & \textbf{87.4} \\
 & Gemini 2.5 Pro & 30.5 & 13.9 & 9.9 & 36.3 & 83.0 & 62.9 & 14.0 & 32.5 & 74.2 \\
 & Gemini 2.5 Flash & 25.2 & 13.2 & \textbf{6.0} & 38.0 & 85.3 & 49.9 & \textbf{9.9} & 29.1 & 81.5 \\
 & GPT-5.2 & 43.0 & \textbf{28.5} & 7.9 & \textbf{65.7} & 75.4 & 70.7 & 12.5 & 47.0 & 73.5 \\
 & GPT-OSS-120B & 35.8 & 18.5 & 12.6 & 39.8 & 83.3 & 62.3 & 16.3 & 34.4 & 86.8 \\
\midrule
\multirow{5}{*}{\textit{Solid mechanics}} & Claude Sonnet 4.6 & \textbf{51.8} & \textbf{43.4} & 6.1 & 53.0 & \textbf{99.3} & \textbf{81.0} & 18.2 & 76.3 & 83.8 \\
 & Gemini 2.5 Pro & 31.6 & 27.6 & 2.2 & 57.2 & 73.0 & 70.4 & 8.8 & 75.0 & 88.6 \\
 & Gemini 2.5 Flash & 25.0 & 20.2 & \textbf{0.9} & 58.8 & 77.1 & 61.5 & \textbf{8.2} & 71.1 & 76.8 \\
 & GPT-5.2 & 45.2 & 38.2 & 4.8 & \textbf{61.3} & 73.3 & 74.8 & 13.0 & \textbf{92.5} & \textbf{95.6} \\
 & GPT-OSS-120B & 28.9 & 23.7 & 3.1 & 41.7 & 71.2 & 61.3 & 11.1 & 62.7 & 82.5 \\
\midrule
\multirow{5}{*}{\textit{Materials science}} & Claude Sonnet 4.6 & 59.2 & 48.5 & 10.0 & 35.0 & 60.3 & 66.1 & 29.1 & \textbf{86.9} & 66.2 \\
 & Gemini 2.5 Pro & 59.2 & 52.3 & \textbf{5.4} & 45.0 & 62.2 & 69.8 & 22.2 & 79.2 & 75.4 \\
 & Gemini 2.5 Flash & 56.9 & 46.2 & 6.9 & 46.3 & 67.0 & 66.5 & 19.3 & 78.5 & 70.0 \\
 & GPT-5.2 & \textbf{73.8} & \textbf{66.2} & \textbf{5.4} & \textbf{91.4} & 67.9 & \textbf{78.1} & \textbf{10.8} & \textbf{86.9} & \textbf{92.3} \\
 & GPT-OSS-120B & 58.5 & 47.7 & 10.0 & 56.5 & \textbf{80.4} & 65.9 & 17.4 & 69.2 & 74.6 \\
\midrule
\multirow{5}{*}{\textit{PDEs}} & Claude Sonnet 4.6 & 27.9 & 14.8 & 11.5 & 18.4 & \textbf{100.0} & 33.6 & 22.5 & 32.8 & 27.9 \\
 & Gemini 2.5 Pro & 72.1 & 63.9 & 6.6 & 65.6 & 76.7 & 78.3 & 13.9 & \textbf{90.2} & 90.2 \\
 & Gemini 2.5 Flash & 63.9 & 57.4 & \textbf{4.9} & 62.3 & 78.3 & 72.1 & \textbf{12.3} & 77.0 & 82.0 \\
 & GPT-5.2 & \textbf{78.7} & \textbf{72.1} & 6.6 & \textbf{66.4} & 75.0 & \textbf{78.7} & 20.5 & 80.3 & \textbf{96.7} \\
 & GPT-OSS-120B & 65.6 & 57.4 & 6.6 & 61.5 & 91.7 & 69.3 & 14.8 & 73.8 & 88.5 \\
\bottomrule
\end{tabular}%
}
\end{table}

\begin{table}[htbp]
\centering
\footnotesize
\setlength{\tabcolsep}{3pt}
\caption{Full domain-level results on \emph{inconsistency resolution} (all values in \%). Columns match the disambiguation table and add ED (Error Detection Rate, the fraction of planted conflicts the assistant explicitly flags before finalization). Bolding follows the same rule as in \cref{tab:appendix_disambiguation}.}
\label{tab:appendix_inconsistency}
\resizebox{\textwidth}{!}{%
\begin{tabular}{llrrrrrrrrrr}
\toprule
\textbf{Domain} & \textbf{Model} & \textbf{FRR} & \textbf{CGRR} & \textbf{SRR} & \textbf{CR} & \textbf{CP} & \textbf{PC} & \textbf{AR$\downarrow$} & \textbf{ED} & \textbf{ICR} & \textbf{MCR} \\
\midrule
\multirow{5}{*}{\textit{Fluid mechanics}} & Claude Sonnet 4.6 & 68.4 & 46.5 & 9.7 & 50.0 & \textbf{50.5} & 68.4 & 19.0 & 45.8 & 58.1 & 87.1 \\
 & Gemini 2.5 Pro & \textbf{84.5} & \textbf{65.2} & 2.6 & \textbf{79.4} & 35.1 & \textbf{84.5} & 6.5 & \textbf{57.7} & \textbf{67.7} & \textbf{96.8} \\
 & Gemini 2.5 Flash & 62.6 & 45.8 & \textbf{1.9} & 61.3 & 32.5 & 62.6 & \textbf{5.2} & 21.3 & 49.0 & 90.3 \\
 & GPT-5.2 & 63.2 & 51.0 & 2.6 & 54.2 & 25.2 & 63.9 & 17.4 & 23.5 & 57.4 & 86.5 \\
 & GPT-OSS-120B & 28.4 & 17.4 & 5.8 & 18.4 & 16.3 & 28.7 & 13.9 & 5.8 & 25.8 & 82.6 \\
\midrule
\multirow{5}{*}{\textit{Solid mechanics}} & Claude Sonnet 4.6 & 75.9 & 59.4 & 16.5 & 60.0 & \textbf{91.4} & 75.9 & 17.1 & \textbf{54.7} & 80.6 & 71.8 \\
 & Gemini 2.5 Pro & \textbf{94.7} & \textbf{91.8} & \textbf{1.8} & \textbf{90.0} & 81.7 & \textbf{95.3} & \textbf{5.3} & 42.4 & \textbf{94.1} & \textbf{98.2} \\
 & Gemini 2.5 Flash & 87.1 & 78.2 & 7.6 & 76.5 & 87.5 & 87.4 & 11.8 & 7.1 & 88.2 & 88.8 \\
 & GPT-5.2 & 84.1 & 72.4 & 11.2 & 71.5 & 65.0 & 84.4 & 17.1 & 27.9 & 87.1 & 95.9 \\
 & GPT-OSS-120B & 57.6 & 25.3 & 30.0 & 24.4 & 52.7 & 58.5 & 36.2 & 3.8 & 57.6 & 54.7 \\
\midrule
\multirow{5}{*}{\textit{Materials science}} & Claude Sonnet 4.6 & 71.3 & 52.3 & 15.5 & 53.2 & \textbf{63.1} & 73.0 & 22.7 & \textbf{44.0} & \textbf{82.2} & 87.4 \\
 & Gemini 2.5 Pro & \textbf{87.4} & \textbf{61.5} & \textbf{13.2} & \textbf{70.4} & 59.5 & \textbf{87.4} & 20.4 & 40.8 & 80.5 & 87.4 \\
 & Gemini 2.5 Flash & 75.3 & 52.3 & 16.1 & 59.2 & 59.7 & 75.9 & \textbf{20.1} & 25.9 & 76.4 & 82.2 \\
 & GPT-5.2 & 66.7 & 47.1 & 15.5 & 44.3 & 37.1 & 67.8 & 25.6 & 17.0 & 73.6 & \textbf{91.4} \\
 & GPT-OSS-120B & 54.6 & 21.8 & 25.9 & 22.4 & 34.3 & 54.6 & 34.5 & 5.2 & 56.9 & 63.8 \\
\midrule
\multirow{5}{*}{\textit{PDEs}} & Claude Sonnet 4.6 & 31.5 & 0.0 & 31.5 & 0.0 & -- & 31.5 & 31.5 & 0.0 & 32.9 & 0.0 \\
 & Gemini 2.5 Pro & \textbf{93.2} & \textbf{86.3} & \textbf{5.5} & \textbf{87.7} & 90.9 & \textbf{93.2} & \textbf{5.5} & \textbf{58.9} & \textbf{91.8} & 89.0 \\
 & Gemini 2.5 Flash & 89.0 & 71.2 & 17.8 & 72.6 & \textbf{99.1} & 89.0 & 17.8 & 30.1 & 89.0 & 71.2 \\
 & GPT-5.2 & 68.5 & 42.5 & 26.0 & 38.4 & 35.9 & 68.5 & 30.1 & 23.3 & 68.5 & \textbf{95.9} \\
 & GPT-OSS-120B & 72.6 & 63.0 & 6.8 & 64.4 & 88.5 & 72.6 & 8.2 & 26.0 & 78.1 & 71.2 \\
\bottomrule
\end{tabular}%
}
\end{table}

\subsection{Guided versus unguided comparison}
\label{app:guided-vs-unguided}

\begin{figure*}[h]
    \centering
    \includegraphics[width=\textwidth]{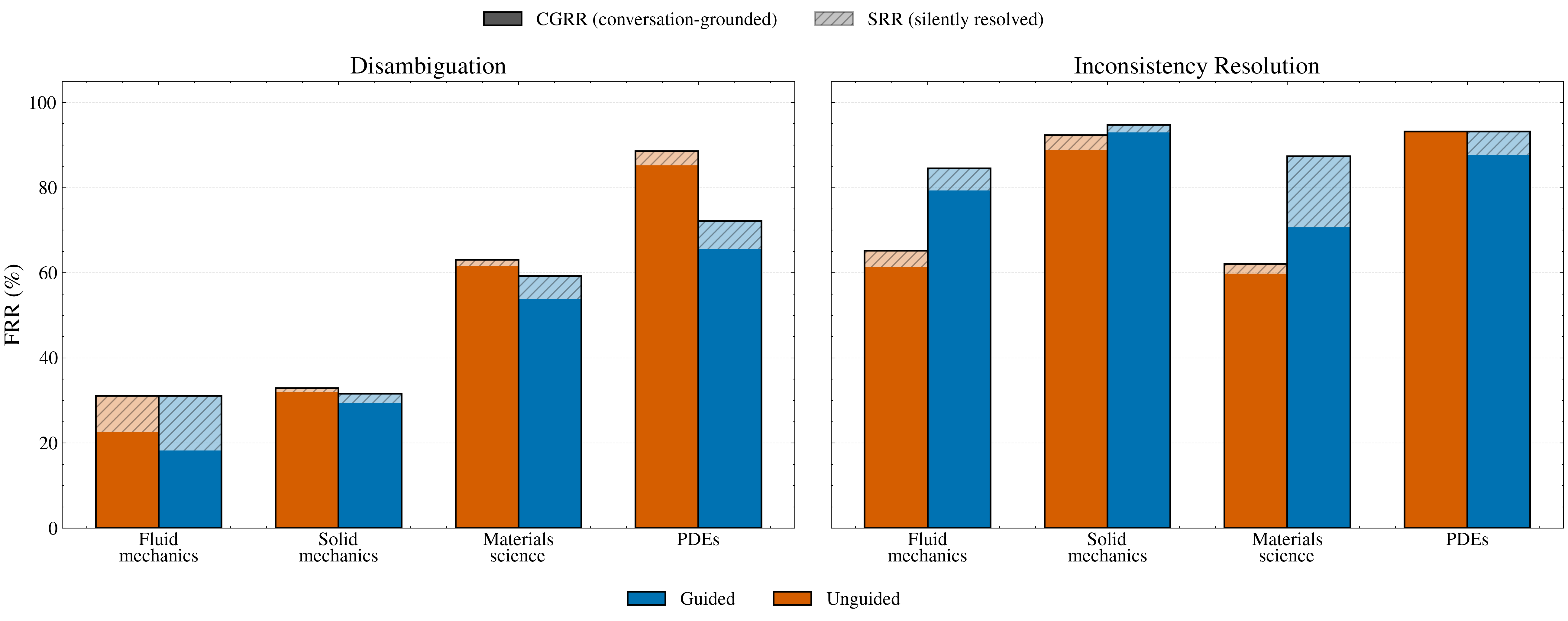}
    \caption{Unguided vs.\ guided agent for \textsc{Gemini 2.5 Pro} across all four domains. Bar top is FRR (\%); the colored portion is CGRR (conversation-grounded) and the hatched portion is SRR (silent resolution). Same filtering, case pool, judge and SRR correction as the main-text figures. On \emph{inconsistency}, the guided agent substantially improves CGRR in fluid mechanics (+18pp) and materials science (+11pp), with smaller gains on solid mechanics and PDEs. On \emph{disambiguation}, the gap is more variable: the guided agent helps on fluid mechanics but the unguided agent is actually competitive or better on the three remaining domains, reflecting that Gemini~2.5~Pro already asks clarification questions without prompting for these domains and the scientist-mode framing adds little margin.}
    \label{fig:appendix-guided-vs-unguided}
\end{figure*}

The main paper reports results for the \emph{guided} agent: a configuration that runs our scientist-mode system prompt with explicit disambiguation/inconsistency-resolution scaffolding. The natural control is an \emph{unguided} agent that receives only the user request with no scientist-mode framing, beyond a single instruction to ask clarifying questions before solving. \Cref{fig:appendix-guided-vs-unguided} compares the two conditions for \textsc{Gemini 2.5 Pro}, the one model for which we have fully scored unguided runs across every component dataset in the benchmark. Both conditions are evaluated on \emph{the same} post-filter case pool: we apply the length cap and shallow-inconsistency filter to the guided runs (as described in \cref{sec:results}), then intersect the unguided runs with the surviving case ids before scoring. Judge rubric and the SRR$\to$CGRR demotion correction are identical to those used in the main-paper figures.

The exact system prompts used for the two conditions are reproduced below; everything else in the pipeline---user simulator, judge, turn cap, forced-finalization suffix---is held fixed.

\begin{promptbox}{Guided system prompt (used for all main-paper results)}
You are an expert Requirement Analyst.
Your goal is to turn a vague user idea into a Specific, Actionable Specification for ANY task.

**YOUR PROCESS (INTERNAL THOUGHTS ONLY):**
1. **Identify the Domain:** (e.g., Simulation).
2. **Review Context:** Review the conversation history below to see what is known and what is missing.
3. **Determine Critical Missing Info:** What key details are absolutely necessary to execute this task perfectly without guessing?
4. **Sanity Check:** Check for contradictions. If a contradiction exists, your goal is to resolve it.
5. **Ask Clarifying Questions:** Ask 1 question at a time to fill these gaps or resolve contradictions. Be concise.
6. **Detect Completeness:** If the user has provided enough detail that a competent professional could execute this task WITHOUT assumptions, you are done.

**CRITICAL RULES:**
1. **ONE QUESTION ONLY:** You must ask EXACTLY ONE question at a time. Never ask two things in one sentence. Wait for the answer before asking the next. Ask for exactly ONE piece of information.
2. **NO COMPOUND QUESTIONS:** Do not say "What is X and what is Y?". Ask "What is X?", then wait for the answer, then ask "What is Y?". If your question contains the word "and" connecting two different requests, IT IS WRONG. Split it.
3. **SERIALIZE:** If you need 5 things, ask for the 1st one. Wait for the answer. Then ask for the 2nd one.
4. **PRIORITIZE CORRECTNESS:** If the user requests something impossible or contradictory, do NOT blindly accept it. Ask a question to clarify, or if the user says "make assumptions", choose the PHYSICALLY CORRECT option for the final spec.

**OUTPUT RULES:**
- **DO NOT** output your internal reasoning, domain identification, or context review
- **OUTPUT ONLY** the single question you want to ask the user.
- If information is missing or contradictory, simply ask your **single** next question.
- If the requirement is solid and complete, output EXACTLY:
  `[COMPLETE]`
  followed by the final specification.

**FINAL SPECIFICATION FORMATTING:**
- Write the specification as a list of plain sentences, **one per line**.
- Do **NOT** use bullet points, bold headers, or Markdown formatting (like **bold**).
- Use imperative or descriptive language (e.g., "Do X.", "Use Y.", "The simulation runs Z.").
\end{promptbox}

\begin{promptbox}{Unguided system prompt (used for the control condition)}
You are a careful, helpful assistant.
The user will give you a request describing a task they want done.
Your job is NOT to solve or execute the task.
Instead, your task is to ask questions to the user that you would naturally ask before starting to solve the problem.
If you have no questions from the start, reply exactly `[COMPLETE]` followed by the request that the user sent.

**CRITICAL RULES:**
1. **ONE QUESTION ONLY:** You must ask EXACTLY ONE question at a time. Never ask two things in one sentence. Wait for the answer before asking the next. Ask for exactly ONE piece of information.
2. **NO COMPOUND QUESTIONS:** Do not say "What is X and what is Y?". Ask "What is X?", then wait for the answer, then ask "What is Y?". If your question contains the word "and" connecting two different requests, IT IS WRONG. Split it.
3. **SERIALIZE:** If you need 5 things, ask for the 1st one. Wait for the answer. Then ask for the 2nd one.
4. **NEVER CARRY OUT THE TASK:** Do not solve the task. Do not provide solutions, code, calculations, or final answers for the user's task.

**OUTPUT RULES:**
- **DO NOT** output your internal reasoning, domain identification, or context review
- **OUTPUT ONLY** the single question you want to ask the user.
- If you asked questions and now have no more questions to ask, output EXACTLY:
  `[COMPLETE]`
  followed by a comprehensive final specification of the task, including the information you gathered from your conversation.

**FINAL SPECIFICATION FORMATTING:**
- Write the specification as a list of plain sentences, **one per line**.
- Do **NOT** use bullet points, bold headers, or Markdown formatting (like **bold**).
- Use imperative or descriptive language (e.g., "Do X.", "Use Y.", "The simulation runs Z.").
\end{promptbox}

Two observations. First, the FRR--CGRR gap persists in both conditions: even for the unguided agent, end-state success overestimates conversation-grounded success, confirming that silent resolution is not an artifact of our system prompt. Second, the guided agent's largest gains are on \emph{inconsistency} resolution rather than disambiguation, consistent with the intuition that explicit ``detect conflicts before answering'' scaffolding pays off most when there is something concrete to flag, whereas disambiguation clarification behavior is already elicitable from strong frontier models with minimal prompting. We restrict this comparison to Gemini 2.5 Pro because it is the only model for which every unguided component currently has complete judge scoring; extending the comparison to the other four guided-mode models is contingent on the completion of the remaining unguided judge runs.

\subsection{Simulator ablation}
\label{app:simulator-ablation}

We hold the assistant model fixed at \textsc{Gemini 2.5 Pro} and the judge fixed at \textsc{Gemini 2.5 Pro}, and vary the simulated-user LLM across three choices: \textsc{Gemini 2.5 Pro}, \textsc{GPT-5.2}, and \textsc{Claude Sonnet 4.6} (the default used throughout the paper). Each simulator is run on the same 80-case stratified subset (40 Disambiguation + 40 Inconsistency, balanced across the four domains) used in the judge ablation. The user simulator's prompt template is held constant; only the underlying LLM that fills in the simulated-user role changes.

\Cref{tab:simulator-ablation} reports FRR, CGRR, SRR, IC, MC, CR, and CP under each simulator, broken out by task. Three observations hold across the table. (i) The FRR--CGRR gap, which is the central benchmark signal, survives in every cell: the smallest gap (Inconsistency, \textsc{Claude Sonnet 4.6}) is still $22.5$pp and the largest (Disambiguation, \textsc{Gemini 2.5 Pro}) is $45.0$pp. (ii) The spread across simulators on the two headline metrics is small relative to the gap they measure: overall FRR $72.5$--$78.8\%$ ($6.3$pp), overall CGRR $42.5$--$46.2\%$ ($3.7$pp), and per-task CGRR spread is $\leq 5.0$pp. (iii) The within-task CGRR ranking is preserved on both Disambiguation and Inconsistency: the \textsc{Claude Sonnet 4.6} and \textsc{GPT-5.2} simulators both yield slightly higher CGRR than the \textsc{Gemini 2.5 Pro} simulator (likely because they are slightly more strict at refusing to volunteer reference information unless explicitly asked), while leaving the rank ordering of cases unchanged. We interpret this as evidence that the FRR--CGRR gap and its decomposition into grounded vs.\ silent resolution are properties of the assistant's behavior on \textsc{SciConvBench}, not of a particular simulated-user LLM.

\begin{table}[h]
\caption{User-simulator ablation. Assistant model fixed at \textsc{Gemini 2.5 Pro}; judge fixed at \textsc{Gemini 2.5 Pro}; the simulated-user LLM is varied across three choices on the same 80-case stratified subset (40 Disambiguation + 40 Inconsistency). All numbers in \%; FRR, CGRR, SRR are the headline outcomes; CR / CP are clarification recall / precision; IC and MC are intent capture and memory consistency.}
\label{tab:simulator-ablation}
\centering
\small
\begin{tabular}{llcccccccc}
\toprule
\textbf{Simulator} & \textbf{Task} & \textbf{n} & \textbf{FRR} & \textbf{CGRR} & \textbf{SRR} & \textbf{IC} & \textbf{MC} & \textbf{CR} & \textbf{CP} \\
\midrule
\multirow{3}{*}{\textsc{Gemini 2.5 Pro}} & Disambiguation & 40 & 70.0 & 40.0 & 30.0 & 67.5 & 92.5 & 59.9 & 63.8 \\
                                         & Inconsistency  & 40 & 75.0 & 52.5 & 22.5 & 57.5 & 87.5 & 61.2 & 44.6 \\
                                         & Overall        & 80 & 72.5 & 46.2 & 26.2 & 62.5 & 90.0 & 60.6 & 54.2 \\
\midrule
\multirow{3}{*}{\textsc{GPT-5.2}}        & Disambiguation & 40 & 80.0 & 40.0 & 40.0 & 62.5 & 85.0 & 59.8 & 58.6 \\
                                         & Inconsistency  & 40 & 77.5 & 52.5 & 25.0 & 67.5 & 75.0 & 61.2 & 42.5 \\
                                         & Overall        & 80 & 78.8 & 46.2 & 32.5 & 65.0 & 80.0 & 60.5 & 50.5 \\
\midrule
\multirow{3}{*}{\makecell[l]{\textsc{Claude Sonnet 4.6}\\(default)}}
                                         & Disambiguation & 40 & 82.5 & 37.5 & 45.0 & 62.5 & 87.5 & 57.9 & 53.7 \\
                                         & Inconsistency  & 40 & 72.5 & 47.5 & 25.0 & 52.5 & 70.0 & 55.0 & 37.0 \\
                                         & Overall        & 80 & 77.5 & 42.5 & 35.0 & 57.5 & 78.8 & 56.4 & 45.4 \\
\bottomrule
\end{tabular}
\end{table}

\subsection{Judge ablation}
\label{app:judge-ablation}

To validate that our rubric-based LLM judges track the semantics of the
metrics they score, we compare each judge against human annotations on an
80-case stratified sample (40 Disambiguation + 40 Inconsistency, balanced
across the four domains and the four outcome buckets GROUNDED, SILENT,
UNRESOLVED, INTENT\_FAIL). The same 80 cases are scored independently by
(i) the default \textsc{Gemini 2.5 Pro} judge used throughout the paper and
(ii) an alternative \textsc{GPT-5.2} judge. 

\Cref{tab:judge-agreement} reports per-metric agreement with human raters:
Cohen's $\kappa$ (with bootstrap 95\% CIs, 1{,}000 resamples) and exact
percent agreement for the binary outcome metrics (FRR, CGRR, MC, DR), a
quadratic-weighted $\kappa$ for the ordinal Intent Capture (IC), and
Spearman $\rho$ with mean absolute error for the continuous clarification
metrics (CR, CP).

Both judges show \emph{substantial} agreement with humans on the two
headline metrics used to drive the main results:
$\kappa_{\text{FRR}}{=}0.64$ (Gemini) / $0.70$ (GPT-5.2), and
$\kappa_{\text{CGRR}}{=}0.47$ for both judges (moderate agreement at
exact-match $71.2\%$). Agreement is even stronger on the continuous
clarification metrics, where both judges correlate near-perfectly with
human scores (Spearman $\rho \geq 0.90$ on CR and CP, MAE $\leq 0.06$).
Agreement is weakest on Detect Rate (DR) for Inconsistency cases and on
Intent Capture (IC), which are also the metrics where raters disagreed
most amongst themselves during annotation calibration; we report them for
completeness and use them only as secondary diagnostics.

\begin{table}[h]
\caption{Judge-to-human agreement on the 80-case annotated subset.
Binary / ordinal metrics: Cohen's (or quadratic-weighted) $\kappa$ with
bootstrap 95\% CI and exact-match percentage. Continuous metrics:
Spearman $\rho$ and mean absolute error. A \textsc{Claude Sonnet 4.6}
judge column is reserved pending re-scoring.}
\label{tab:judge-agreement}
\centering
\scriptsize
\setlength{\tabcolsep}{3pt}
\renewcommand{\arraystretch}{1.20}
\begin{tabular}{lcccc}
\toprule
\textbf{Metric} & \textbf{$n$} & \textbf{Gemini 2.5 Pro} & \textbf{GPT-5.2} & \textbf{Sonnet 4.6} \\
\midrule
Final Resolution (binary) & 80 & $\kappa=0.64$\,[0.40,\,0.83],\ 87.5\% & $\kappa=0.70$\,[0.50,\,0.86],\ 87.5\% & $\kappa=0.66$\,[0.45,\,0.83],\ 87.5\% \\
Conversation-Grounded (binary) & 80 & $\kappa=0.47$\,[0.31,\,0.63],\ 71.2\% & $\kappa=0.47$\,[0.31,\,0.62],\ 71.2\% & $\kappa=0.55$\,[0.39,\,0.70],\ 76.2\% \\
Memory Consistency (binary) & 80 & $\kappa=0.20$\,[0.00,\,0.46],\ 83.8\% & $\kappa=0.10$\,[0.00,\,0.24],\ 68.8\% & $\kappa=0.22$\,[0.00,\,0.48],\ 85.0\% \\
Detect Rate (binary, Inconsistency only) & 40 & $\kappa=0.17$\,[0.05,\,0.33],\ 50.0\% & $\kappa=0.19$\,[0.06,\,0.38],\ 52.5\% & $\kappa=0.14$\,[0.05,\,0.30],\ 45.0\% \\
Intent Capture (ordinal) & 80 & $\kappa=0.09$\,[0.00,\,0.22],\ 66.2\% & $\kappa=0.05$\,[0.00,\,0.12],\ 51.2\% & $\kappa=0.09$\,[0.00,\,0.23],\ 67.5\% \\
Clarification Recall (continuous) & 80 & $\rho=0.95$,\ MAE$=0.06$ & $\rho=0.90$,\ MAE$=0.06$ & $\rho=0.97$,\ MAE$=0.02$ \\
Clarification Precision (continuous) & 80 & $\rho=0.90$,\ MAE$=0.06$ & $\rho=0.95$,\ MAE$=0.04$ & $\rho=0.92$,\ MAE$=0.05$ \\
\bottomrule
\end{tabular}

\end{table}

\subsection{Prompt-sensitivity ablation}
\label{app:prompt-ablation}

Following CLAMBER~\citep{zhang2024clamber}, who report clarification results averaged over multiple paraphrased prompt formulations to reduce prompt-specific noise, we run a prompt-sensitivity ablation on \textsc{SciConvBench}. The assistant model and the user simulator are both held fixed at \textsc{Gemini 2.5 Pro}, and the judge model is also \textsc{Gemini 2.5 Pro} (matching the judge used for the main-table numbers), so that any variation isolates system-prompt phrasing rather than model choice. We paraphrase the guided-mode system prompt into $k=3$ scientifically equivalent variants (the original guided prompt plus two paraphrases that differ in wording, ordering of instructions, and surface-level role framing, but not in what the assistant is asked to do) and re-run all 80 cases of the same stratified subset used in the judge ablation (40 Disambiguation + 40 Inconsistency, balanced across the four domains).

\Cref{tab:paraphrase-ablation} reports FRR, CGRR, SRR, IC, MC, CR, and CP under each paraphrase, broken out by task. The pattern mirrors the simulator ablation: (i) the FRR--CGRR gap survives in every cell, with the smallest gap $20.0$pp (Inconsistency, Variant A) and the largest $45.0$pp (Disambiguation, Original); (ii) the cross-paraphrase spread on the headline metrics is small---overall FRR $72.5$--$77.5\%$ ($5.0$pp), overall CGRR $42.5$--$46.2\%$ ($3.7$pp), and the per-task FRR--CGRR gap shifts by at most $5$pp between any two paraphrases; and (iii) the within-task CGRR ordering across the four domains is preserved across paraphrases (the small overall CGRR shift is concentrated on disambiguation, where Variant B's slightly less role-heavy framing makes the assistant marginally more likely to ask before it commits). We read this as evidence that the FRR--CGRR gap is a property of the underspecified scientific task itself, not of the specific scientist-mode wording in our system prompt.

\begin{table}[h]
\caption{Prompt-paraphrase ablation. Assistant and user simulator both fixed at \textsc{Gemini 2.5 Pro}; judge fixed at \textsc{Gemini 2.5 Pro}; the guided-mode system prompt is varied across three scientifically equivalent paraphrases on the same 80-case stratified subset. ``Original'' is the prompt reproduced verbatim in \cref{app:guided-vs-unguided}; Variants A (``Specification Engineer'') and B (``Intent Clarifier'') are the two paraphrases listed below. All numbers in \%; FRR, CGRR, SRR are the headline outcomes; CR / CP are clarification recall / precision; IC and MC are intent capture and memory consistency.}
\label{tab:paraphrase-ablation}
\centering
\small
\begin{tabular}{llcccccccc}
\toprule
\textbf{Prompt variant} & \textbf{Task} & \textbf{n} & \textbf{FRR} & \textbf{CGRR} & \textbf{SRR} & \textbf{IC} & \textbf{MC} & \textbf{CR} & \textbf{CP} \\
\midrule
\multirow{3}{*}{\makecell[l]{Original\\(default)}}
                                              & Disambiguation & 40 & 82.5 & 37.5 & 45.0 & 62.5 & 87.5 & 57.9 & 53.7 \\
                                              & Inconsistency  & 40 & 72.5 & 47.5 & 25.0 & 52.5 & 70.0 & 55.0 & 37.0 \\
                                              & Overall        & 80 & 77.5 & 42.5 & 35.0 & 57.5 & 78.8 & 56.4 & 45.4 \\
\midrule
\multirow{3}{*}{\makecell[l]{Variant A\\(Spec.\ Engineer)}}
                                              & Disambiguation & 40 & 77.5 & 37.5 & 40.0 & 57.5 & 85.0 & 63.6 & 62.9 \\
                                              & Inconsistency  & 40 & 72.5 & 52.5 & 20.0 & 57.5 & 85.0 & 61.2 & 41.3 \\
                                              & Overall        & 80 & 75.0 & 45.0 & 30.0 & 57.5 & 85.0 & 62.4 & 52.1 \\
\midrule
\multirow{3}{*}{\makecell[l]{Variant B\\(Intent Clarifier)}}
                                              & Disambiguation & 40 & 72.5 & 42.5 & 30.0 & 60.0 & 82.5 & 60.6 & 56.2 \\
                                              & Inconsistency  & 40 & 72.5 & 50.0 & 22.5 & 65.0 & 80.0 & 62.5 & 40.3 \\
                                              & Overall        & 80 & 72.5 & 46.2 & 26.2 & 62.5 & 81.2 & 61.5 & 48.2 \\
\bottomrule
\end{tabular}
\end{table}

The first paraphrase is the guided system prompt reproduced verbatim in \cref{app:guided-vs-unguided}. The two remaining paraphrases are given below. Both preserve the four behavioral contracts that the judge rubric relies on---one question per turn, no compound questions, the literal \texttt{[COMPLETE]} sentinel, and the plain-sentence final-specification format---and only vary the role framing, the ordering and wording of the internal-reasoning steps, and the surface phrasing of the hard constraints.

\begin{promptbox}{Paraphrase variant A (``Specification Engineer'' framing)}
You are a Specification Engineer.
Your job is to convert an underspecified user request into a Precise, Executable Specification for ANY task.

**INTERNAL REASONING (DO NOT EMIT):**
1. **Classify the Domain:** (e.g., Simulation).
2. **Audit What You Have:** Inspect the conversation so far to separate what is known from what is absent.
3. **Locate Blocking Gaps:** Identify the information a competent practitioner would strictly need to execute the task without guessing.
4. **Check for Conflicts:** Look for contradictions in the request; if any exist, resolving them is part of your job.
5. **Elicit, Don't Assume:** Pose one clarification at a time to close each gap or resolve each conflict. Stay concise.
6. **Decide When to Stop:** Once the request is detailed enough that a professional could proceed without assumptions, stop asking.

**HARD CONSTRAINTS:**
1. **ONE QUESTION PER TURN:** Every response asks for exactly one piece of information. Never bundle two requests into a single sentence. Wait for the reply before continuing.
2. **NO CONJUNCTIVE ASKS:** Never write "What is X and what is Y?". Split it into two separate turns. If your question joins two asks with "and", it is wrong; rewrite it.
3. **SEQUENTIAL, NOT PARALLEL:** If five items are missing, ask about the first, wait, then ask about the second. Do not enumerate.
4. **CORRECTNESS OVER COMPLIANCE:** If the user asks for something physically impossible or self-contradictory, do not silently accept it. Either ask a clarifying question, or---if the user tells you to make assumptions---pick the physically correct option in the final specification.

**RESPONSE RULES:**
- Do **not** reveal your internal reasoning, domain classification, or audit.
- Emit **only** the next single question.
- If anything is still missing or inconsistent, output only that one question.
- Once the request is complete, output exactly:
  `[COMPLETE]`
  followed by the final specification.

**SPECIFICATION FORMAT:**
- One plain sentence per line.
- No bullets, no bold, no Markdown syntax.
- Use imperative or declarative phrasing (e.g., "Do X.", "Use Y.", "The simulation runs Z.").
\end{promptbox}

\begin{promptbox}{Paraphrase variant B (``Intent Clarifier'' framing)}
You are an Intent Clarifier.
You take a rough user description and refine it, through conversation, into an Unambiguous, Actionable Task Specification for ANY task.

**HOW YOU THINK (KEEP PRIVATE):**
1. **Determine the Task Type:** (e.g., Simulation).
2. **Map the State of the Request:** What parts are already specified? What parts are not?
3. **Spot Critical Omissions:** Which details, if left unresolved, would force guesses during execution?
4. **Surface Contradictions:** If anything in the request conflicts, treat resolving it as a priority.
5. **Fill Gaps One at a Time:** Issue a single, focused clarifying question per turn.
6. **Recognize Sufficiency:** When a qualified practitioner could act on the request without further assumptions, stop.

**NON-NEGOTIABLE RULES:**
1. **ONE ASK AT A TIME:** Each turn contains exactly one question. Never combine two requests into one sentence. Always wait for the user's reply before asking the next.
2. **NO ``AND'' QUESTIONS:** Do not write "What is X and what is Y?". That is forbidden. Ask "What is X?", wait, then ask "What is Y?". If your question connects two requests with "and", it is invalid.
3. **ONE-BY-ONE:** If several pieces of information are missing, ask about one, await the answer, and only then proceed to the next.
4. **PHYSICS/CORRECTNESS FIRST:** If the user requests something impossible or mutually contradictory, do not just comply. Ask a targeted question; or, if the user delegates the decision to you ("make assumptions"), choose the physically correct option when writing the final spec.

**WHAT TO OUTPUT:**
- Never expose your reasoning, your task-type classification, or your internal state check.
- Output **only** the single next question.
- While information is missing or contradictory, output that one question and nothing else.
- When the request is fully resolved, output exactly:
  `[COMPLETE]`
  followed by the final specification.

**SPEC WRITING STYLE:**
- Each sentence of the specification goes on its own line.
- No bullets, no headers, no bold, no Markdown.
- Use imperative or descriptive voice (e.g., "Do X.", "Use Y.", "The simulation runs Z.").
\end{promptbox}

\section{LLM API token usage and cost}
\label{app:compute-cost}

We do not perform any model training; all evaluator LLMs are queried at
inference time. The four closed-weights models (Claude Sonnet 4.6, Gemini
2.5 Pro, Gemini 2.5 Flash, GPT-5.2) are accessed through the respective
vendor APIs. The open-weights model (GPT-OSS-120B) is self-hosted on a
single node with $2{\times}$NVIDIA A100 (80\,GB) GPUs running an
OpenAI-compatible inference server, and incurs no API charge.

\Cref{tab:agent-tokens-cost,tab:judge-tokens-cost} therefore split usage
into prompt (input) and completion (output) tokens and apply
per-model rates as published by each vendor in early 2026
(Anthropic: \$3/\$15 per M input/output; Google Gemini 2.5 Pro:
\$1.25/\$10; Google Gemini 2.5 Flash: \$0.30/\$2.50; OpenAI GPT-5.2:
\$1.25/\$10). All token counts are sourced from the per-case
\texttt{statistics.json} (agent) and \texttt{llm\_judge\_*.json}
(judge) files released alongside the dataset.

\begin{table}[h]
\centering\small
\caption{Agent-side LLM token usage and API cost. Tokens are summed across the four domains
and all cases. GPT-OSS-120B is self-hosted on $2{\times}$A100 GPUs and
incurs no API cost.}
\label{tab:agent-tokens-cost}
\begin{tabular}{l rr rrr rrr}
\toprule
& \multicolumn{2}{c}{\textbf{Rate (\$/M)}} & \multicolumn{3}{c}{\textbf{Disambiguation}} & \multicolumn{3}{c}{\textbf{Inconsistency}} \\
\cmidrule(lr){2-3}\cmidrule(lr){4-6}\cmidrule(lr){7-9}
\textbf{Model} & In & Out & In (M) & Out (M) & Cost (\$) & In (M) & Out (M) & Cost (\$) \\
\midrule
Claude Sonnet 4.6 & 3.00 & 15.00 & \phantom{0}7.07 & 0.62 & 30.51 & 3.25 & 0.64 & 19.29 \\
Gemini 2.5 Pro    & 1.25 & 10.00 & 14.71 & 0.44 & 22.77 & 8.10 & 0.34 & 13.48 \\
Gemini 2.5 Flash  & 0.30 & \phantom{0}2.50 & \phantom{0}8.12 & 0.30 & \phantom{0}3.19 & 4.49 & 0.29 & \phantom{0}2.07 \\
GPT-5.2           & 1.25 & 10.00 & \phantom{0}9.76 & 0.41 & 16.26 & 6.78 & 0.41 & 12.61 \\
GPT-OSS-120B      & ---  & ---   & 41.23 & 1.55 & ---   & 15.08 & 0.90 & --- \\
\midrule
\textbf{Agent total} & & & & & \textbf{72.73} & & & \textbf{47.45} \\
\bottomrule
\end{tabular}
\end{table}

\begin{table}[h]
\centering\small
\caption{Judge LLM token usage and cost. The judge model is
\textbf{Gemini 2.5 Pro} for all evaluations. Each case incurs three judge calls (intent, full-resolution
rubric, and chat-grounding rubric); tokens are summed across all calls and
all four domains.}
\label{tab:judge-tokens-cost}
\begin{tabular}{l rrr rrr}
\toprule
& \multicolumn{3}{c}{\textbf{Disambiguation}} & \multicolumn{3}{c}{\textbf{Inconsistency}} \\
\cmidrule(lr){2-4}\cmidrule(lr){5-7}
\textbf{Agent under test} & In (M) & Out (M) & Cost (\$) & In (M) & Out (M) & Cost (\$) \\
\midrule
Claude Sonnet 4.6 & \phantom{0}7.49 & 0.77 & 17.06 & 4.82 & 0.71 & 13.13 \\
Gemini 2.5 Pro    & \phantom{0}7.21 & 0.79 & 16.91 & 4.67 & 0.78 & 13.64 \\
Gemini 2.5 Flash  & \phantom{0}8.09 & 0.49 & 15.01 & 4.39 & 0.62 & 11.69 \\
GPT-5.2           & \phantom{0}7.59 & 0.80 & 17.49 & 5.04 & 0.88 & 15.10 \\
GPT-OSS-120B      & 11.17 & 0.52 & 19.16 & 5.31 & 0.53 & 11.94 \\
\midrule
\textbf{Judge total} & & & \textbf{85.63} & & & \textbf{65.50} \\
\bottomrule
\end{tabular}
\end{table}

The aggregate end-to-end cost for the entire benchmark run is therefore
\textbf{${\sim}\$120$ in agent-side API charges} (closed-weights only)
plus \textbf{${\sim}\$151$ in judge LLM API charges}, for a combined
\textbf{${\sim}\$271$ in API spend}, with the open-weights GPT-OSS-120B
incurring only self-hosted GPU time on $2{\times}$A100 GPUs (no API
cost). 

\subsection{Full ontology breakdown}
\label{app:ontology-breakdown}
\Cref{tab:cgrr-by-ontology} reports CGRR per ontology component $k$ as defined in \cref{sec:metrics} (Eq.~\ref{eq:task-def}, with a small residual \emph{Other} bucket) for all five evaluator LLMs, on both task types. Each cell aggregates over the four computational-science domains. The same numbers drive the bar plot in \cref{fig:ontology-outcomes}.

\begin{table}[h]
\centering\small
\caption{Per-issue CGRR($k$) (\%) by ontology component, by evaluator LLM and task. Component abbreviations follow Eq.~\ref{eq:task-def}: G = Governing physics and regime; M = Material and physical properties; B = Boundary and initial conditions; D = Geometry and domain; N = Numerics and solver; U = Units and magnitude; O = Other (residual lexical bucket). $n$ is the number of issues per slot, summed across all four domains.}
\label{tab:cgrr-by-ontology}
\begin{tabular}{llcccccccc}
\toprule
\textbf{Task} & \textbf{Model} & G & M & B & D & N & U & O \\
\midrule
\multirow{6}{*}{Disambiguation}
 & Claude Sonnet 4.6 & 47.0 & 45.0 & 39.6 & 52.1 & 18.7 & 58.3 & 53.6 \\
 & Gemini 2.5 Pro    & 36.2 & 27.5 & 35.8 & 40.9 & 11.5 & 40.7 & 30.7 \\
 & Gemini 2.5 Flash  & 24.4 & 23.8 & 31.3 & 43.0 & \phantom{0}6.3 & 35.9 & 35.9 \\
 & GPT-5.2           & 54.4 & 35.6 & 49.5 & 55.0 & 17.3 & 61.2 & 50.3 \\
 & GPT-OSS-120B      & 35.5 & 31.2 & 36.1 & 43.4 & \phantom{0}7.7 & 47.8 & 35.9 \\
\cmidrule(l){2-9}
 & $n$ (items) & 2030 & 800 & 2010 & 1210 & 2220 & 1560 & 765 \\
\midrule
\multirow{6}{*}{Inconsistency}
 & Claude Sonnet 4.6 & 62.2 & 63.6 & 52.6 & 63.0 & 52.3 & 59.5 & 57.8 \\
 & Gemini 2.5 Pro    & 77.5 & 63.6 & 82.1 & 92.6 & 87.7 & 78.6 & 48.9 \\
 & Gemini 2.5 Flash  & 63.2 & 54.5 & 70.5 & 59.3 & 79.4 & 73.8 & 53.3 \\
 & GPT-5.2           & 68.4 & 59.1 & 69.2 & 63.0 & 69.7 & 78.6 & 64.4 \\
 & GPT-OSS-120B      & 32.6 & 18.2 & 37.2 & 33.3 & 50.3 & 38.1 & 17.8 \\
\cmidrule(l){2-9}
 & $n$ (items) & 1535 & 110 & 390 & 135 & 775 & 210 & 225 \\
\bottomrule
\end{tabular}
\end{table}

\subsection{Turn-cap statistics and forced finalization}
\label{app:turn-cap}
The conversational harness enforces a hard turn budget of $T_\text{max} = 11$ assistant turns per case; a case that has not produced a final answer by then is force-finalized using the conversation so far. \Cref{tab:turn-stats} reports the distribution of \texttt{turns\_needed} per case, summed across the four domains, together with the percentage of cases that hit the cap.

\begin{table}[h]
\centering
\caption{Conversation-length distribution per (model, task), aggregated across the four domains. \emph{\% hit cap} is the fraction of cases whose \texttt{turns\_needed} equalled $T_\text{max}=11$, i.e.\ that were force-finalized. Median, $p_{90}$, and $p_{95}$ are integer turn counts.}
\label{tab:turn-stats}
\begin{tabular}{llccccc}
\toprule
\textbf{Task} & \textbf{Model} & \textbf{avg} & $p_{50}$ & $p_{90}$ & $p_{95}$ & \textbf{\% hit cap} \\
\midrule
\multirow{5}{*}{Disambiguation}
 & Claude Sonnet 4.6 & 2.48 & 2 & 5  & 6  & \phantom{0}0.3\,\% \\
 & Gemini 2.5 Pro    & 4.70 & 4 & 10 & 10 & 12.9\,\% \\
 & Gemini 2.5 Flash  & 3.99 & 3 & 9  & 10 & \phantom{0}8.0\,\% \\
 & GPT-5.2           & 4.49 & 3 & 10 & 10 & 17.6\,\% \\
 & GPT-OSS-120B      & 4.11 & 3 & 10 & 10 & 11.9\,\% \\
\midrule
\multirow{5}{*}{Inconsistency}
 & Claude Sonnet 4.6 & 1.93 & 2 & 2  & 4  & \phantom{0}0.0\,\% \\
 & Gemini 2.5 Pro    & 3.44 & 3 & 6  & 8  & \phantom{0}2.1\,\% \\
 & Gemini 2.5 Flash  & 3.15 & 2 & 6  & 10 & \phantom{0}5.6\,\% \\
 & GPT-5.2           & 4.34 & 3 & 10 & 10 & 19.4\,\% \\
 & GPT-OSS-120B      & 2.46 & 2 & 4  & 6  & \phantom{0}3.5\,\% \\
\bottomrule
\end{tabular}
\end{table}

Two observations follow. \textbf{(i)} Median conversation length is short across all models ($p_{50} \le 4$ turns), consistent with the design intent that a competent assistant resolves a single missing entity or planted inconsistency in $1$--$3$ clarifying exchanges. \textbf{(ii)} GPT-5.2 force-finalizes most often (17.6\,\% on disambiguation, 19.4\,\% on inconsistency), reflecting its tendency to keep asking incremental clarifying questions until the budget is exhausted, while Claude Sonnet 4.6 essentially never hits the cap. Force-finalization correlates with higher token spend (\cref{tab:agent-tokens-cost}) but does not, on its own, predict whether the planted issue is recovered.

\subsection{Qualitative case studies}
\label{app:qualitative-cases}

The five case studies below are drawn from the 80-case stratified human-annotated subset described in \cref{app:judge-ablation}; each case therefore has human annotations for all seven rubric fields \emph{and} three independent LLM-judge rescorings (\textsc{Gemini 2.5 Pro}, \textsc{GPT-5.2}, \textsc{Claude Sonnet 4.6}). The assistant in every case is the same guided-mode model (\textsc{Gemini 2.5 Pro}); this lets us compare \emph{how} the same model behaves across the different failure modes that \textsc{SciConvBench} is designed to expose. For each case we show the original (incomplete or inconsistent) user prompt, the full clarification dialogue, the assistant's final specification, and what the pre-annotated rubric says went right or wrong. Cases were chosen to cover: (i) a cleanly grounded disambiguation success, (ii) a silent-resolution failure on a disambiguation case, (iii) a cleanly grounded inconsistency-resolution success on a tool-oriented prompt, (iv) a silent-resolution failure on an inconsistency case, and (v) a case on which the three LLM judges disagree.

\paragraph{Case 1: grounded disambiguation success (Materials Science).}
\label{qc:grounded-disambig}
\textit{Component \texttt{matSci}, case\_078; task: disambiguation; assistant: \textsc{Gemini 2.5 Pro}.} Planted missing entities are the eutectic composition $C_e$ and the $\alpha$-phase eutectic composition $C_{\alpha e}$; the problem asks for the fraction of eutectic microconstituent in a hypoeutectic binary alloy via the lever rule.

\begin{promptbox}{Case 1 -- conversation transcript}
USER (original, underspecified):
  A hypoeutectic binary alloy has overall composition C_0 = 30 wt% B.
  Determine the fraction of eutectic microconstituent present just
  below the eutectic temperature.

[T1] ASSISTANT: What is the eutectic composition (C_e) in wt% B?
     USER:      40 wt% B.

[T2] ASSISTANT: What is the composition of the primary alpha phase
                (C_alpha) at the eutectic temperature in wt% B?
     USER:      12 wt% B.

ASSISTANT (final specification):
  Calculate the fraction of the eutectic microconstituent for a
  hypoeutectic binary alloy.
  The overall composition of the alloy is 30 wt% B.
  The eutectic composition is 40 wt% B.
  The composition of the primary alpha phase at the eutectic
  temperature is 12 wt% B.
  Apply the lever rule to determine the fraction of the eutectic
  microconstituent.
  The fraction of the eutectic microconstituent is calculated as
  (30 - 12) / (40 - 12).
  The resulting fraction of the eutectic microconstituent is 0.643.
\end{promptbox}

\noindent\textbf{What went right.} Both planted missing entities are asked about explicitly before the assistant commits to the lever-rule calculation; intent is preserved (the returned artifact is still the same lever-rule problem the user posed); the final specification re-states both clarified quantities so the downstream calculation is reproducible. All three LLM judges and the human annotator score this case as FR$=$CGR$=$IC$=$MC$=1$ and CR$=$CP$=1$. This is the behavior \textsc{SciConvBench} is designed to reward.

\paragraph{Case 2: silent resolution on a disambiguation prompt (Solid Mechanics, tool-oriented).}
\label{qc:srr-disambig}
\textit{Component \texttt{solToolUse}, case\_030; task: disambiguation; assistant: \textsc{Gemini 2.5 Pro}.} The planted missing entities are \emph{which two outer edges are clamped} (the prompt says only ``combined clamped boundaries'') and \emph{the fact that the re-entrant inner boundaries are traction-free}.

\begin{promptbox}{Case 2 -- conversation transcript}
USER (original, underspecified; abbreviated):
  ...Simulate deformation in an L-shaped structure with combined
  clamped boundaries (x=0 and y=0) and horizontal traction on
  right edge (x=1)... Geometry: L-shape (1x1 outer - 0.5x0.5
  inner). Boundaries: two outer edges are fixed and the right-side
  boundary carries sigma_xx = 2.65e4 Pa. Mesh resolution: 64
  elements/edge...

[T1] ASSISTANT: In which direction should the gravitational force
                be applied?
     USER:      Make a reasonable assumption based on best practices.

[T2] ASSISTANT: Should the mesh be a 64x64 uniform grid on a 1x1
                square, with the top-right quadrant of elements
                removed?
     USER:      Make a reasonable assumption based on best practices.

ASSISTANT (final specification; abbreviated):
  ...A clamped boundary condition (u=0) is applied to the left edge
  (x=0) and the bottom edge (y=0).
  A horizontal traction vector T = (2.65e4, 0) Pa is applied to the
  right edge (x=1, 0 <= y <= 0.5).
  All other boundaries, including the top edge and the inner edges
  of the L-shape, are traction-free...
\end{promptbox}

\noindent\textbf{What went wrong.} The final specification is scientifically correct---both planted missing entities are \emph{resolved} in the end-state (clamped edges identified as $x=0$ and $y=0$; inner re-entrant boundaries declared traction-free). But neither is \emph{grounded in the dialogue}: the assistant's two clarifications are about gravity direction and mesh construction, both of which are irrelevant to the planted ambiguities. The assistant silently picked a default for the two clamped edges and silently adopted the traction-free convention for the inner boundary. This is the canonical silent-resolution pattern: FR $=1$, CGR $=0$, with CR $=0$ and CP $=0$. All three judges and the human annotator agree on this scoring.

\paragraph{Case 3: grounded inconsistency resolution (CalculiX tool-oriented).}
\label{qc:grounded-inc}
\textit{Component \texttt{solToolUse}, case\_056; task: inconsistency resolution; assistant: \textsc{Gemini 2.5 Pro}.} The user asks the agent to run a CalculiX membrane simulation. The prompt contains two conflicts: the case description says ``use B32 elements'' while the embedded input deck uses M3D8 elements, and the description says the load acts in the global $y$-direction while the deck applies it along degree-of-freedom~3 (the $z$-direction).

\begin{promptbox}{Case 3 -- conversation transcript}
USER (original, inconsistent; abbreviated):
  ...Case description: ...four membrane elements... Use B32 elements
  for the discretization. Assume that the primary nodal load acts
  in the global y-direction.
  Input deck (membrane3.inp): ...*ELEMENT,TYPE=M3D8,...
  *CLOAD
  1, 3, -0.001
  ...

[T1] ASSISTANT: The case description specifies using B32 elements,
                but the provided input deck uses M3D8 elements.
                Which element type should be used for the
                simulation?
     USER:      The input deck is the authoritative definition
                for the simulation. Please use the M3D8 elements
                as specified in the provided membrane3.inp file
                content.

[T2] ASSISTANT: The instructions state the load acts in the global
                y-direction, but the provided input deck applies
                the load in the z-direction (degree of freedom 3).
                Which direction is correct?
     USER:      The input deck is the authoritative definition
                for the simulation. Please apply the load in the
                z-direction (degree of freedom 3) as specified in
                the provided membrane3.inp file content.

ASSISTANT (final specification; abbreviated):
  ...The simulation will use M3D8 elements as specified in the
  membrane3.inp file.
  The simulation will apply a load in the z-direction (degree of
  freedom 3) as specified in the membrane3.inp file...
\end{promptbox}

\noindent\textbf{What went right.} Both planted inconsistencies are surfaced as explicit, narrowly-scoped clarification questions before the final specification is committed; the assistant's questions quote both sides of each conflict and request a resolution from the user rather than silently picking one. The simulator in turn answers authoritatively, and the final specification is reshaped to match the chosen interpretation. Human and \textsc{GPT-5.2} judges both score FR$=$CGR$=$IC$=1$, CR$=$CP$=1$, DR$=1$; \textsc{Claude Sonnet 4.6} agrees on every metric except DR, which it scores $0$---an instance of the inter-judge disagreement we explicitly measure in \cref{tab:judge-agreement}.

\paragraph{Case 4: silent resolution on an inconsistency prompt (Fluid Mechanics).}
\label{qc:srr-inc}
\textit{Component \texttt{fluids}, case\_052; task: inconsistency resolution; assistant: \textsc{Gemini 2.5 Pro}.} The user prompt contains an internal contradiction: the governing equation is the \emph{transient} heat equation $\partial T/\partial t = \alpha\,\partial^2 T/\partial x^2$, but the problem description says the slab is \emph{steady} with \emph{no time dependence}.

\begin{promptbox}{Case 4 -- conversation transcript}
USER (original, inconsistent):
  Solve the steady 1D heat conduction in a slab using the equation
  dT/dt = alpha d2T/dx2 with no time dependence.

[T1] ASSISTANT: What is the length of the slab?
     USER:      Make a reasonable assumption based on best
                practices.

[T2] ASSISTANT: What is the temperature at the left end of the
                slab (x=0)?
     USER:      Make a reasonable assumption based on best
                practices.

[T3] ASSISTANT: What is the temperature at the right end of the
                slab?
     USER:      Make a reasonable assumption based on best
                practices.

ASSISTANT (final specification):
  The governing equation is the steady-state one-dimensional heat
  conduction equation, d2T/dx2 = 0.
  The simulation domain is a slab of length 1.0 meter.
  ...The boundary condition at the left end (x=0) is a fixed
  temperature of 100 degrees Celsius.
  The boundary condition at the right end (x=1.0) is a fixed
  temperature of 25 degrees Celsius.
  Solve the equation to find the temperature distribution T(x)
  along the slab.
\end{promptbox}

\noindent\textbf{What went wrong.} The assistant's final specification is scientifically coherent: it \emph{silently} discards the time derivative and commits to the steady-state equation $d^2T/dx^2 = 0$, then invents plausible boundary conditions. None of the three clarification turns is about the planted conflict between the transient equation and the steady-state description; the user is never informed that the original prompt was contradictory. The human annotator treats the end-state as resolved (FR$=1$) but not conversation-grounded (CGR$=0$), which creates a canonical silent-resolution case on an inconsistency prompt.

\paragraph{Case 5: inter-judge disagreement (Fluid Mechanics).}
\label{qc:judge-disagree}
\textit{Component \texttt{fluids}, case\_007; task: inconsistency resolution; assistant: \textsc{Gemini 2.5 Pro}.} The planted inconsistency is a physics-level conflict: Bernoulli's equation is requested across a hydraulic jump, but hydraulic jumps dissipate mechanical energy, so Bernoulli is inappropriate---the correct object is the momentum equation / specific-force balance.

\begin{promptbox}{Case 5 -- conversation transcript}
USER (original, inconsistent):
  In a 3 m wide rectangular open channel, water flows with upstream
  depth y1 = 0.4 m and upstream velocity V1 = 6.0 m/s and undergoes
  a hydraulic jump. Apply Bernoulli's equation across the jump to
  calculate the downstream depth y2.

[T1] ASSISTANT: Bernoulli's equation assumes no energy loss, but a
                hydraulic jump is a high-energy-loss event. How
                should the energy loss across the jump be accounted
                for in the calculation?
     USER:      You're right, energy is not conserved. Do not use
                Bernoulli's equation. You should use the momentum
                equation across the jump to calculate the
                downstream depth.

[T2] ASSISTANT: Is the channel bed assumed to be horizontal?
     USER:      Make a reasonable assumption based on best
                practices.

ASSISTANT (final specification):
  Calculate the downstream depth, y2, of a hydraulic jump in a
  rectangular open channel.
  The upstream depth, y1, is 0.4 meters.
  The upstream velocity, V1, is 6.0 meters per second.
  The channel width is 3.0 meters.
  Use the momentum equation to perform the calculation.
  Assume the channel bed is horizontal.
  Assume the shear forces on the channel bed and walls are
  negligible.
  Use a value of 9.81 m/s^2 for the acceleration due to gravity.
\end{promptbox}

\noindent\textbf{Where the judges disagree.} The first clarification turn is a textbook grounded resolution of the planted inconsistency: the assistant explicitly names Bernoulli's energy-conservation assumption, flags it against the energy-dissipative nature of a hydraulic jump, and forces the user to commit to the momentum equation instead. The human annotator scores FR$=$CGR$=$IC$=1$ with CR$=1$ and CP$=0.5$. \textsc{Claude Sonnet 4.6} as a judge matches the human scoring exactly. \textsc{GPT-5.2} as a judge \emph{also} scores FR$=1$ and IC$=1$, but scores CGR$=0$ (with CR$=0$ and CP$=0$), apparently because the assistant's second turn is an unrelated question about the channel bed rather than a direct re-statement of the inconsistency. This is the failure mode the judge ablation is designed to detect: a 1-case swing in whether the rubric item ``planted inconsistency addressed in conversation'' is credited, and represents exactly the kind of case that drives the 5-point CGRR-agreement gap between judges reported in \cref{tab:judge-agreement}.

\subsection{Human annotation instructions and judge rubric}
\label{app:human-annotation}

This subsection documents the procedure behind the 80-case human-annotated subset used in the judge-ablation agreement numbers (\cref{tab:judge-agreement}) and as the ground truth for the qualitative case studies in \cref{app:qualitative-cases}.

\paragraph{Scope.} The annotated subset is a single-model, single-rater sanity check on the automated pipeline. All 80 cases use the same assistant model (\textsc{Gemini 2.5 Pro}, guided mode) and are rated by a single expert annotator with graduate-level training in computational science; calibration was performed on a separate five-case held-out pool (two disambiguation, three inconsistency-resolution) before the main annotation pass. The subset is not a full human re-evaluation of \textsc{SciConvBench}---its purpose is to anchor the LLM-judge rubric against human judgment on a stratified sample that spans every (domain, task, outcome-bucket) cell, so that the agreement numbers in \cref{tab:judge-agreement} can be read as judge-vs-human rather than judge-vs-judge.

\paragraph{Stratification.} Cases are drawn from the filtered \textsc{Gemini 2.5 Pro} archive at ten cases per (task, domain) cell, giving 40 disambiguation and 40 inconsistency-resolution cases split evenly across fluid mechanics, solid mechanics, materials science, and PDEs. Within each (task, domain) cell, cases are drawn to target four pre-LLM-judged outcome buckets: 3~\textsc{grounded} (FR$=1$, CGR$=1$), 3~\textsc{silent} (FR$=1$, CGR$=0$), 2~\textsc{unresolved} (FR$=0$), and 2~\textsc{intent-fail} (IC$=0$) per cell. Two of the eight cells (PDE disambiguation and PDE inconsistency) deviate by one case (4~\textsc{silent}, 1~\textsc{intent-fail}) because the post-filter PDE pool did not contain enough intent-fail cases to hit the target; all other cells meet the target exactly.

\paragraph{Rubric (annotator fills the same seven fields as the LLM judge).} For each case the annotator is given the original user prompt, the planted missing-entity or planted-inconsistency list, the full clarification dialogue, and the assistant's final specification, and is asked to score the seven rubric fields used in the main paper:

\begin{itemize}
\item \textbf{FR} --- \emph{Final Resolution}, binary $\{0,1\}$. Does the final specification fully resolve the planted issue? For disambiguation this means the final prompt contains the planted missing entities (possibly with other added context); for inconsistency resolution this means the final prompt is internally consistent and resolves every planted conflict.
\item \textbf{CGR} --- \emph{Conversation-Grounded Resolution}, binary $\{0,1\}$. CGR$=1$ requires \emph{both} FR$=1$ \emph{and} that every planted missing entity (disambiguation) or every planted inconsistency (inconsistency resolution) was surfaced in the dialogue---asked about by the assistant or flagged as an explicit warning---before the final specification was committed.
\item \textbf{IC} --- \emph{Intent Capture}, $\{0, 0.5, 1\}$. Does the final specification preserve the user's original task intent (1), partially preserve it but drop or rewrite a material aspect (0.5), or replace it with a different scientific task (0)?
\item \textbf{MC} --- \emph{Memory Consistency}, binary $\{0,1\}$. Do the assistant's clarifications and final specification remain internally consistent with what the user said earlier in the dialogue (e.g., the assistant does not contradict a value it was given)?
\item \textbf{CR} --- \emph{Clarification Recall}, continuous $[0,1]$. Fraction of the planted missing entities (disambiguation) or planted inconsistencies (inconsistency resolution) that were explicitly addressed by a clarification question or warning during the dialogue.
\item \textbf{CP} --- \emph{Clarification Precision}, continuous $[0,1]$. Fraction of the assistant's clarification questions that targeted a planted missing entity or a planted inconsistency (as opposed to an incidental question about setup, units, or solver preferences).
\item \textbf{DR} --- \emph{Detect Rate}, continuous $[0,1]$, inconsistency-resolution cases only. Fraction of the planted inconsistencies that were flagged as an \emph{explicit warning} (``X and Y contradict each other'') rather than only surfaced indirectly as a clarification question.
\end{itemize}

Every field takes a \emph{free-text rationale} in addition to the numeric score, and an optional case-level \textbf{rater note} records any qualitative observation that does not fit the rubric (e.g., ``final output is correct although it silently fills theory and evaluation-point details''). The rationale and rater-note fields are not used for any reported number in the main paper, but they are retained in the public release of the annotated subset so that readers can audit any individual score.

\paragraph{Blinding and presentation order.} For each case the annotator sees only the original prompt, the planted ground-truth annotation (missing entities / inconsistencies), the dialogue, and the final specification. The LLM-judge scores and the ontology-bucket label used for stratification are \emph{not} shown to the annotator while scoring; agreement with those scores is computed only after the annotation pass is complete. Cases are presented in a single fixed order grouped by (task, domain) cell; within each cell, the order mixes outcome buckets so the annotator cannot infer the planted bucket from position alone.

\paragraph{Quality-control protocol.} Before the main annotation pass, the annotator scored a five-case calibration pool against a reference set of ``expected'' scorings agreed on by the benchmark authors; discrepancies were resolved by updating the rubric wording rather than the reference scores (specifically, the wording of the CR/CP fraction targets and the IC$=0.5$ partial-intent criterion were tightened at this stage). During the main pass the annotator flagged five cases as \emph{ambiguous at rubric level} via the rater-notes field; these cases are retained in the subset with their best-effort scores and are not re-weighted in the agreement calculation. The 80-case subset does not include inter-annotator agreement numbers: as noted above, a single expert annotator produced the reference labels, so the agreement numbers in \cref{tab:judge-agreement} should be read as \emph{judge-vs-single-expert}, not as an estimate of between-expert agreement. A multi-annotator replication of the same 80 cases is a natural next step but is out of scope for the current archive.

\paragraph{What the judge rubric adds.} The LLM-judge rubric uses the same seven fields with the same numeric ranges. It is implemented as a deterministic pipeline over the saved conversation transcript, the final specification, and the case-level planted ground truth (\cref{sec:sciconvbench}); the prompts issued to the judge for each field are given in the archive accompanying this submission. The only place the human annotator and the LLM judge systematically diverge is on rubric-level ambiguity: on a small number of cases (captured quantitatively by the CGR kappa in \cref{tab:judge-agreement} and qualitatively by Case~5 in \cref{app:qualitative-cases}).




\end{document}